\documentclass[twocolumn,letterpaper]{IEEEAerospaceCLS}  % only supports two-column, letterpaper format

\usepackage[]{graphicx}    % We use this package in this document
\usepackage[colorlinks=true, urlcolor=blue, linkcolor=black, citecolor=black]{hyperref} % ED 2025 added for web links
\newcommand{\ignore}[1]{}  % {} empty inside = %% comment
\usepackage{xcolor}
\usepackage{ifthen}
\usepackage{cite}
\usepackage{amsmath}
\usepackage{amssymb}
\usepackage{amstext}
\usepackage{amsfonts}
\usepackage{booktabs}
\usepackage{multirow}
\usepackage{stfloats}
\usepackage{makecell}

\usepackage[nameinlink]{cleveref}
\Crefname{section}{Section}{Sections}
\Crefname{subsection}{Subsection}{Subsections}

\newboolean{comparemode}
\setboolean{comparemode}{false}

\newcommand{\hr}[1]{%
  \ifthenelse{\boolean{comparemode}}%
    {\textcolor{red}{#1}}%
    {#1}%
}

\crefname{appendix}{Appendix}{Appendices}
\Crefname{appendix}{Appendix}{Appendices}

\begin{document}
\title{Monocular Navigation Relative to Unknown Spacecraft Using a Transformer-Aided Kalman Filter}

\author{%
Pol Francesch Huc\\ 
Department of Aeronautics and Astronautics\\
Stanford University\\
Stanford, CA 94305\\
pfrances@stanford.edu
\and 
Simone D'Amico\\
Department of Aeronautics and Astronautics\\
Stanford University\\
Stanford, CA 94305\\
damicos@stanford.edu
%%%% IMPORTANT: Use the correct copyright information--IEEE, Crown, or U.S. government. %%%%%
\thanks{\footnotesize 979-8-3195-1048-8/27/$\$31.00$ \copyright2027 IEEE} 
}

\maketitle

\thispagestyle{plain}
\pagestyle{plain}

\begin{abstract}
As space becomes increasingly congested and contested, robust autonomous navigation is essential for servicing, debris removal, and situational awareness with respect to unknown resident space objects.
This work presents a novel learning-based pipeline for pose estimation of uncooperative and unknown spacecraft using only monocular images from a single servicer.
The approach combines a transformer-based neural network with a Multi-State Constraint Kalman Filter (MSCKF) to estimate the pose (i.e., position and orientation of the target spacecraft relative to the camera) throughout rendezvous and proximity operations.
Unlike existing vision-based methods that require prior knowledge of the target shape or inertia properties, or rely on additional sensing modalities such as depth, lidar, or stereo, the proposed pipeline generalizes to previously unseen spacecraft using a single monocular camera. 
The transformer network estimates the odometry, the change in pose between images up to scale, from SuperPoint features matched by LightGlue. 
These odometry estimates serve as MSCKF pseudo-measurements.

The MSCKF uses these pseudo-measurements along with an orbit and attitude kinematics model to estimate the pose of the target.
In particular, the relative orbit elements, the target's attitude with respect to the servicer's camera (where the target's body is defined arbitrarily by the filter), and the associated angular velocity are estimated directly by the filter.
\hr{Because the camera is monocular, the range is not estimated: the filter tracks the relative orbit up to scale, and a separate study shows the range becomes observable with known orbit maneuvers.}
Feature tracks and loop closures are used to bound short- and long-term drift, respectively. 
The filter is initialized using coarse relative orbit information that could be available from upstream methods such as angles-only navigation, and no other prior such as attitude, shape, size or inertia properties.
The method is trained and evaluated on a re-rendered high-resolution version of the SPE3R dataset, which includes synthetic images of 103 spacecraft. 
Eleven of these spacecraft are held out during training to evaluate the generalization to unseen targets.
Monte Carlo simulations are then used to evaluate the navigation pipeline on rendered trajectories of the held out spacecraft.
\hr{The network is further evaluated on the SHIRT trajectories with rendered and hardware-in-the-loop images}, where it retains a $4$--$6$ fold advantage in the median translation direction error over a classical baseline. 
The results demonstrate that learned vision pipelines as a front-end for Kalman filters provide median errors of \hr{$2.5^\circ$ in attitude and $0.9\%$ of range in the scale-normalized ROE} when navigating about unknown targets.
The approach supports inspection, servicing, and debris removal missions involving targets without available shape models. 
\end{abstract} 

\vspace{-1.0cm}
\tableofcontents

%%%%%%%%%%%%%%%%%%%%%%%%%%%%%%%%%%%%%%
\section{Introduction}
%%%%%%%%%%%%%%%%%%%%%%%%%%%%%%%%%%%%%%
\begin{figure*}[t]
    \includegraphics[width=\textwidth]{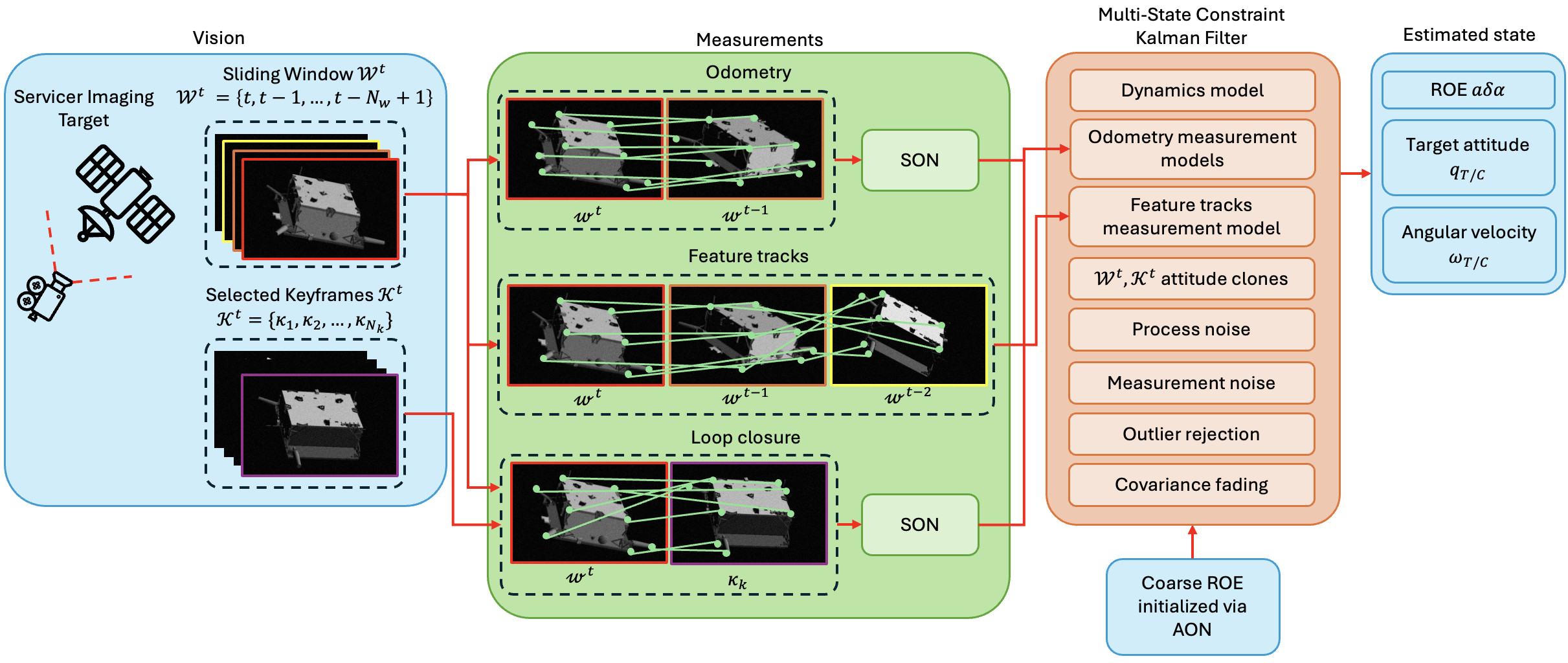}
    \centering
    \caption{Proposed pipeline for pose estimation where a servicer captures images of a target, and uses those images to generate three measurements. These measurements leverage feature extraction and matching to estimate odometry between the current image and the previous one, feature tracks along the sliding window, and loop closure between the current image and candidates in the selected keyframes. The odometry is estimated using the novel Spacecraft Odometry Network (SON). These three measurements provide short- and long-baseline relative information that is leveraged by a Multi-State Constraint Kalman Filter that then tracks the desired pose states. For every image in the sliding window and for every keyframe, the filter maintains a clone of the target attitude estimate. \vspace{-0.5cm}}
    \label{fig:overall_pipeline}
\end{figure*}

The space industry is experiencing a resurgence in interest, especially in the realm of autonomous rendezvous, proximity operations, and docking (RPOD). 
These missions generally require satellites, denoted as servicers, to be close to their target with separations on the order of meters to contact. 
However, the target may not always be cooperative (because it is defunct, not designed for RPOD, or adversarial) and so the satellite is required to independently perform pose estimation of the target to successfully complete its mission. 
Unknown or damaged targets may also require autonomous shape reconstruction.
Cameras are an attractive choice for this kind of mission since they are generally cheap, lightweight, power-efficient, and already ubiquitous on satellites as they are used in attitude estimation.

The approach taken in this work is to first estimate the pose of the target with respect to the servicer, without having access to the 3D model.
Pose estimates support safe operation during image collection and provide inputs for concurrent shape estimation.
The pose of the target is estimated from an incoming stream of images taken by the servicer during close range operations after being initialized by an earlier phase of the mission such as an approach using Angles-Only Navigation (AON) \cite{kruger2024starlingformationflyingopticalexperiment}.

Although pose estimation of known targets through monocular images is a mature technology \cite{d_pose_2014}, the same cannot be said of unknown targets. 
This work combines a novel odometry network with a  Multi-State Constraint Kalman Filter (MSCKF) \cite{mourikis2007multi} for monocular pose estimation of unknown targets. 
A notional pipeline is shown in \autoref{fig:overall_pipeline}. 

Because monocular measurements are scale-free, the range \hr{is unobservable} when no prior on the target's size or range is available. 
\hr{The filter therefore estimates the relative orbit up to scale, and \autoref{sec:scale_observability} studies how known maneuvers make the range observable.}
As images are collected they are immediately added to the sliding window $\mathcal{W}^t$ which stores the most recent images in a cache. 
Images which provide large pose diversity are stored in a cache of selected keyframes $\mathcal{K}^t$, and are replaced only by images that increase that diversity. 

From these images, features are extracted by SuperPoint \cite{detone_superpoint_2018}.
Since the target 3D model is unknown, regressing the pose of the target from a single set of features or image is difficult, as is seen by the poor pose estimates in the work by Park et al. \cite{park_2024_scitech_spe3r}.
Instead, this work estimates features in sets of images, and uses the state-of-the-art feature matcher LightGlue \cite{lindenberger_lightglue_2023} to match the features between the images. 
LightGlue is fine-tuned on the novel dataset SPE3R+ to adapt it to spacecraft imagery.

SPE3R+ builds upon the original SPE3R \cite{park_2024_scitech_spe3r} dataset, which included 1,000 images and masks per model for 64 \hr{unit-scale} satellite models, with 7 of those models reserved for a test set (so they were never seen during training). 
SPE3R+ includes these 64 original models, but adds 39 new models (103 total, 11 in test set), \hr{makes them real-scale}, and provides 5,000 higher resolution images per model, \hr{along} with masks, and metric depth.

These feature matches on their own are not sufficient to estimate the pose, and typical computer vision pipelines will use the 8-point algorithm along with RANdom SAmple Consensus (RANSAC), or similar, to estimate the fundamental matrix and the odometry between pairs of images \cite{hartley_multiple_2004}. 
As will be shown, the estimation accuracy from this method is poor, and so the Spacecraft Odometry Network (SON) is introduced to estimate the odometry from noisy feature matches. 

Because SON only provides an odometry measurement, subsequent filters are prone to divergence since this is a form of dead reckoning. 
Two additional measurements are used to prevent this divergence: feature tracks and loop closure. 
Feature tracks are features observed across multiple images in the sliding window and provide multi-frame relative constraints that limit local drift.
Loop closure provides a longer-baseline relative attitude constraint between the current image and a retained keyframe.

An MSCKF is then introduced to estimate the pose of the target over a trajectory. 
The MSCKF is used to leverage the feature tracks directly, as these provide the geometric constraint needed for this kind of filter without needing to incorporate the feature locations on the target in the filter state. 
Outlier rejection and covariance inflation are also leveraged to remove potentially noisy feature track and loop closure measurements as well as reduce the overconfidence in the relative orbit estimate. 

This work contributes:
\begin{itemize}
    \item An expanded and re-rendered dataset of 103 satellite models with grayscale images, masks, and depths, SPE3R+ \footnote{Available from: https://purl.stanford.edu/rn464ph6457}.
    \item \hr{A novel network for odometry estimation from noisy feature matches, SON, enabling vision-odometry for previously unseen spacecraft under large illumination changes.}
    \item \hr{A novel implementation of the MSCKF, leveraging SON, feature tracks and loop closures instead of IMUs to accurately estimate the relative pose from a trajectory of incoming images.}
    \item \hr{Demonstration of pose estimation with range using orbit maneuvers for an unknown target for the first time, to the authors' best knowledge.}
\end{itemize}

This paper begins by discussing the state of the art in this field in \Cref{sec:sota}. 
Next, the proposed architecture is discussed in Sections  \ref{sec:feature_extraction_matching}--\ref{sec:msckf}.
The implementation details are discussed in \Cref{sec:training_and_implementation_details} covering key assumptions, datasets used, and hyperparameter selection. 
\Cref{sec:results} evaluates the components individually where possible, and the complete pipeline.  

%%%%%%%%%%%%%%%%%%%%%%%%%%%%%%%%%%%%%%
\section{State of the art} \label{sec:sota}
%%%%%%%%%%%%%%%%%%%%%%%%%%%%%%%%%%%%%%
The state of the art in monocular pose estimation begins with known targets. 
DARPA's Orbital Express demonstrated cooperative vision-based rendezvous in 2007 using a camera and target-mounted retro-reflectors \cite{4526518}.
D'Amico et al. \cite{d_pose_2014} instead used classical computer vision techniques to detect edge features on the target and then use the known 3D model to regress the 6D pose, demonstrating uncooperative pose estimation. 
Improvements on the pose estimate have been made through the use of Convolutional Neural Networks (CNN) \cite{sharma_pose_2018}, which have more recently been traded for Vision Transformers (ViT) \cite{park2024spnv3} for reduced computational cost and increased robustness to the domain gap.
The process has been shown to be robust to out of distribution images through the implementation of an Unscented Kalman Filter (UKF) \cite{park2023spnukf} and by refining the network on the target domain \cite{park2024spnv2}.
Recent surveys \cite{pauly2023survey, koushafar2026uncertainty} identify deployability on space-grade hardware, further closing of the domain gap, and reliable uncertainty under domain shift as the open problems in this line of work, all of which still assume a known target.

These methods, and others, are trained on a large amount of images and their popularity has led to the release of image datasets of different satellites. 
The dataset by Dung et al. \cite{Dung_2021_CVPR} compiles real spacecraft imagery with bounding-box, segmentation, and component level annotations, but does not provide metric poses. 
URSO \cite{9197244} in contrast uses photorealistic rendering to generate synthetic spacecraft images with pose labels for known target geometries. 
SPEED+ \cite{park2022speedplus}, SHIRT \cite{park2023spnukf}, and MAN-DATA \cite{lebreton_training_2024} include synthetic and real images in a space-surrogate environment and provide reference truth poses, but only of one target each. 
Finally SPE3R \cite{park_2024_scitech_spe3r} is a dataset of synthetic images only, but it includes 64 real satellites in its dataset along with the pose labels. 
Due to the number of targets and the availability of poses SPE3R is most useful for this work, however the images are provided at a very low resolution and so it is improved here. 

Literature on unknown target pose estimation in space is much more limited. 
Park et al. \cite{park_2024_scitech_spe3r} introduce a CNN to estimate the pose and shape from a single image, however the pose estimates have errors in the tens of degrees for the test set, and the network does not estimate the scale. 
Bates et al. \cite{bates2025removing} improve pose estimation by addressing canonical pose ambiguities, but still do not recover the metric scale. 
DreamSat \cite{mathihalli2024dreamsat} and DreamSat-Pose \cite{uwumukiza2026dreamsat} predict shape and, in the latter case, pose, but still do not estimate the scale and provide similar errors as Park et al. \cite{park_2024_scitech_spe3r}.
CRISP \cite{shi_crisp_2024} has better pose estimates than Park et al. \cite{park_2024_scitech_spe3r}, but requires RGB-D images. 
Barad et al. \cite{Barad_2024} also simultaneously estimate the shape and pose, but they do so using a sequence of RGB-D images which provides the scale, and tracking tends to drift over long duration trajectories. 
Dor et al. \cite{dor_orb-slam_2018} apply Simultaneous Localization And Mapping (SLAM) using ORB-SLAM \cite{7219438} to space-like imagery, but validation is limited to 3-DOF motion and tracking is inconsistent. 
Augenstein \cite{augenstein2011monocular} implements a landmark-based SLAM using monocular images only, and recovers both the pose and shape but again does not estimate the scale. 
Capuano et al. \cite{capuano2020monocular} and  Tweddle et al. \cite{tweddle2015factor} both estimate the pose of unknown targets, but they are aided by a laser range finder and stereovision, respectively. 

Two works do provide monocular vision-based pose estimation, but they also include significant drawbacks.
Ticozzi et al. \cite{ticozzi2024factor} implement a factor-graph based active SLAM method which generates an attitude guidance strategy to minimize the target shape model uncertainty. 
Their strategy relies on perfect initialization (the first two poses are assumed to be known) and feature matching is assumed to be perfect as well. 
The method is only validated over a short horizon, at most 23 poses.
Legrand et al. \cite{legrand2024leveraging} follow a two-stage approach, where a NeRF model \cite{mildenhall2021nerf} is used to generate imagery for training a conventional known target pose estimation network \cite{park2024spnv2}. 
However, the first stage requires pose estimates which limits the online applicability.

Terrestrial vision-based pose estimation on the other hand is a much larger field. Algorithms like COLMAP \cite{schoenberger2016mvs, schoenberger2016sfm} estimate the pose and structure from a set of images, but are typically run offline. 
These have been recently complemented by learning-based reconstruction from image pairs, such as DUSt3R \cite{wang2024dust3r} and MASt3R \cite{leroy2024grounding}, which regress dense pointmaps with large transformers on terrestrial scenes.
Feature-based methods use feature matches between two images in order to estimate the odometry between the images, mostly by estimating the fundamental matrix \cite{hartley_multiple_2004}. 
Early feature detectors and matchers include SIFT \cite{lowe_distinctive_2004}, SURF \cite{bay_surf_2006}, and ORB \cite{6126544}, but they are increasingly being replaced by learning based methods in current literature.
SuperPoint \cite{detone_superpoint_2018} supplies learned features for attention-based matchers SuperGlue \cite{sarlin2020superglue} and LightGlue \cite{lindenberger_lightglue_2023}, with the latter reducing inference time through early stopping. 
Other learned approaches include DISK \cite{tyszkiewicz2020disk}, dense matchers such as RoMa \cite{edstedt2024roma}, and simultaneous extraction and matching methods such as 
LoFTR \cite{sun_loftr_2021}. 

Some terrestrial methods focus then on improving the odometry estimation (sometimes with scale) that must occur from these feature matches.
Rockwell et al. \cite{rockwell_8-point_2022} introduce a transformer which regresses the odometry directly from an image pair, and FAR \cite{rockwell2024farflexibleaccuraterobust} extends this by combining such a learned prediction with a match-based solver.
Although FAR supports multiple feature and correspondence front ends, its learned metric-scale prior is trained on terrestrial scenes and does not directly address unknown spacecraft with arbitrary physical scale.
Arnold et al. \cite{arnold2022map} introduce map-free visual relocalization where odometry estimation is done directly from images, but this requires a large training dataset not available for space applications.

Existing work does not yet resolve the problem of accurate monocular pose estimation of unknown targets. 
Solutions which target space-like imagery generally make simplifying assumptions, such as the availability of additional sensors, or \hr{do not track the relative pose and orbit over long trajectories}.
Terrestrial methods on the other hand are not purpose built for space-like imagery, and so do not function out of the box with the same performance as they do on the domain they are designed for.
This work seeks to resolve this by leveraging some terrestrial state-of-the-art methods and adapting them to the space problem, along with a novel learning-based solution to odometry estimation.

%%%%%%%%%%%%%%%%%%%%%%%%%%%%%%%%%%%%%%
\section{Feature Extraction \& Matching on Image Pairs} \label{sec:feature_extraction_matching}
%%%%%%%%%%%%%%%%%%%%%%%%%%%%%%%%%%%%%%

SuperPoint \cite{detone_superpoint_2018} first extracts image keypoints and descriptors for a pair of images, and LightGlue \cite{lindenberger_lightglue_2023} estimates correspondences between them. 
In this work, a keypoint and its descriptor are denoted as a feature.
Although these methods perform well on terrestrial benchmarks \cite{jin2021image}, their performance on spacecraft imagery is poor, and motivates domain adaptation through fine-tuning.
Additionally, images are cropped using a target bounding box (derived from reference truth in this work, see \autoref{sec:training_and_implementation_details}) before feature extraction to remove noisy background features. 

\begin{figure}[htbp]
    \centering
    \includegraphics[width=\linewidth]{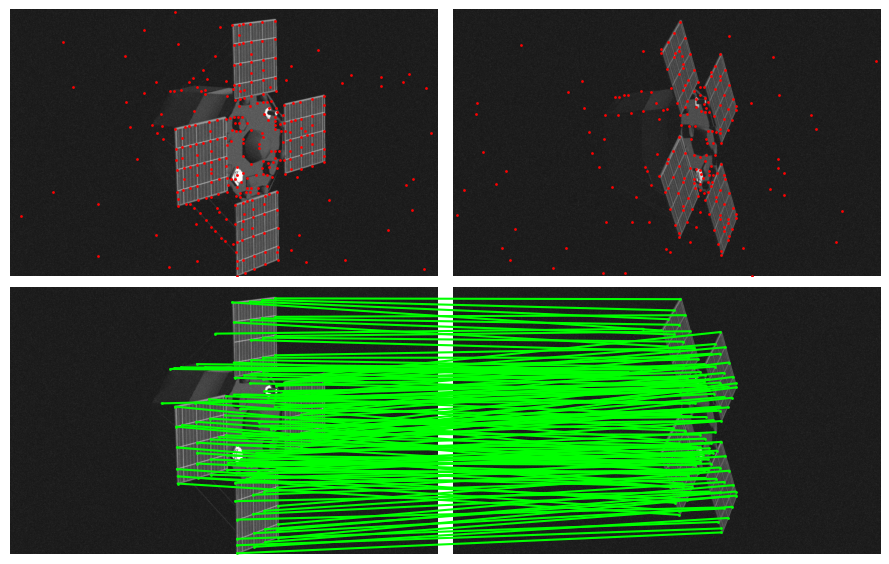}
    \caption{Features extracted by SuperPoint (top) and matched by LightGlue (bottom) from a pair of images of the ACRIMSAT satellite. }
    \label{fig:spe3r_lightglue_matches}
\end{figure}

For the domain adaptation, LightGlue is fine-tuned on SPE3R+ while SuperPoint remains frozen.
Features are therefore generated by SuperPoint during training, but reference truth correspondences are generated by projecting the extracted features between the images using the known reference truth poses and metric depth.
This is followed by nearest-neighbor association with thresholds listed in \autoref{tab:training_parameters} in \autoref{sec:training_and_implementation_details}.
Evaluation uses $11$ spacecraft that are withheld during training, and so represent the true unknown target case.

SuperPoint and LightGlue are chosen because SuperPoint provides a fixed set of features for each image, allowing LightGlue matches to be chained into multi-image feature tracks. 
This property motivates their use over pair-dependent correspondences such as those produced by LoFTR \cite{sun_loftr_2021}.
The resulting tracks supply the MSCKF with the geometric constraints described in \autoref{sec:msckf}.

\autoref{fig:spe3r_lightglue_matches} illustrates the extracted features and matches.

%%%%%%%%%%%%%%%%%%%%%%%%%%%%%%%%%%%%%%
\section{Odometry Estimation from Feature Matches} \label{sec:odometry_estimation_from_feature_matches}
%%%%%%%%%%%%%%%%%%%%%%%%%%%%%%%%%%%%%%
Let $\mathbf{q}_{T/C}^t$ and ${}^C \mathbf{r}_{T/C}^t$ denote the target attitude and position relative to the camera at time $t$, using an arbitrary fixed target frame for the orientation. 
Between images at $t-1$ and $t$ the relative rotation $\Delta \mathbf{q}^{t, t-1}$ and translation $\mathbf{r}^{t, t-1}$ are,
\begin{equation}
    \Delta \mathbf{q}^{t, t-1} = \mathbf{q}_{T/C}^t \otimes \left( \mathbf{q}_{T/C}^{t-1} \right)^* \label{eqn:odometry_quaternion_update}
\end{equation}
\begin{equation}
    \mathbf{r}^{t, t-1} = {}^C\mathbf{r}_{T/C}^t - \mathbf{R}^{t, t-1} {}^C\mathbf{r}_{T/C}^{t-1}
\end{equation}
where $\otimes$ denotes quaternion multiplication, $\mathbf{q}^*$ quaternion conjugation, and $\mathbf{R}^{t,t-1}$ is the rotation matrix representing $\Delta \mathbf{q}^{t, t-1}$. 
Both of the estimators covered in this section however only estimate the relative rotation and the unit translation direction $\hat{\mathbf{u}} = \mathbf{r}^{t, t-1} / \| \mathbf{r}^{t, t-1} \|$, which together are dubbed the odometry. 
Metric scale observability is \hr{discussed in} \autoref{sec:scale_observability}.

\subsection{Classical Approach}
The classical baseline estimates a fundamental matrix from the feature matches using OpenCV's USAC implementation \cite{raguram2012usac} with MAGSAC++ scoring \cite{barath_magsac_2020}.
Known camera intrinsics then convert this to an essential matrix, whose decomposition and cheirality check recover the relative rotation $\Delta \mathbf{q}^{t, t-1}$ and translation direction $\hat{\mathbf{u}}$ \cite{hartley_multiple_2004}. 

\subsection{SON: Spacecraft Odometry Network}
\begin{figure*}[t]
    \centering
    \includegraphics[width=\linewidth]{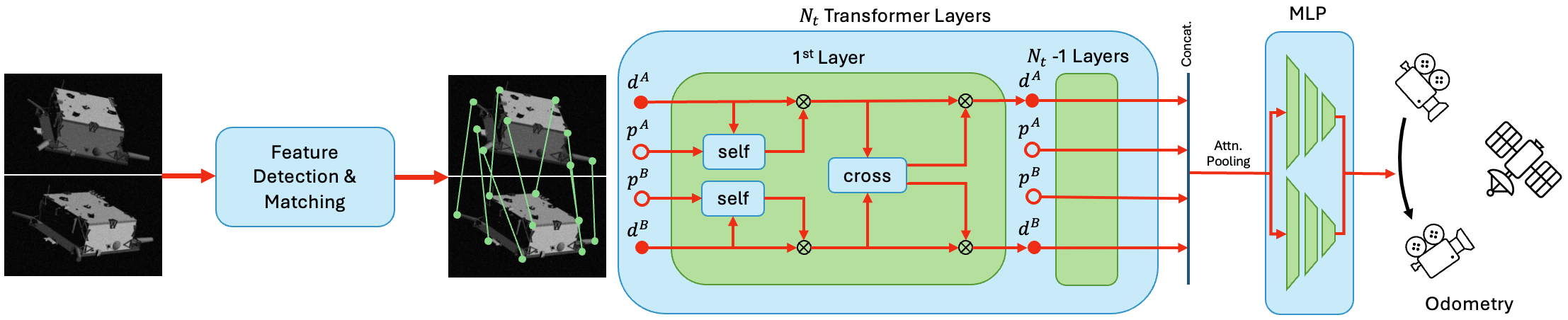}
    \caption{Architecture of SON, where feature estimates from SuperPoint are matched with LightGlue and the descriptors $d$ and their positional encodings $p$ are used to regress the odometry between images $A$ and $B$. \vspace{-0.5cm}}
    \label{fig:son_architecture}
\end{figure*}

SON is a transformer-based neural network architecture designed to regress the odometry between two images from noisy feature matches. 
The architecture is shown in \autoref{fig:son_architecture} and uses the transformer backbone from LightGlue \cite{lindenberger_lightglue_2023}, along with a Multilayer Perceptron (MLP) to regress the odometry. 
To address the fact that LightGlue does not provide a constant number of feature matches, attention-based pooling is performed to select the most important portions of the feature matches and descriptors and provide a constant input size for the MLP. 
The reason for using LightGlue is that feature matching and odometry estimation are coupled problems: given the odometry and depth, matches follow by projection, as in \autoref{sec:feature_extraction_matching}.
The philosophy behind using LightGlue then is that neural networks which perform well in the task of feature matching could also perform well in related tasks. 
The results and the ablation studies performed in this work are used to validate this claim, giving some confidence for the choice in architecture. 

Like the baseline, SON only estimates the odometry up to a scale. 
This design decision is made because doing otherwise would require the network to learn about the size of satellites since the scale of the translation vector is fully coupled with the size of the target, which is more difficult for monocular vision to do while generalizing to unseen targets. 
\hr{The filter therefore tracks the relative orbit up to scale, and \Cref{sec:scale_observability} discusses how the scale can be recovered.} 

\subsubsection{Transformer backbone}
The transformer backbone is made up of $N_t$ identical transformer layers \cite{vaswani2017attention}, each of which take as input the descriptors $d \in \mathbb{R}^{N_m \times 256}$ and the position encodings $p \in \mathbb{R}^{N_m \times 128}$ of the feature matches for both images, where $N_m$ is the number of feature matches. 
The transformer then outputs the same variables, but is trained to add context to the descriptors with self- and cross-attention blocks which is helpful for the downstream task of odometry estimation. 
Like LightGlue, SON uses a learnable Fourier encoder for the positional encoding \cite{li2021learnable, su2024roformer}.

For particular details on the transformer architecture readers are referred to LightGlue \cite{lindenberger_lightglue_2023}, but a short summary is provided here. 
For each layer, each set of descriptors and matches from an image passes through the self-attention block where the descriptors are augmented using information from other descriptors and matches in the same image. 
The cross-attention block then takes as input the descriptors and positional encodings from both images, and uses information across the images to augment the descriptors.

The feature locations are first mapped to 2D bearing coordinates through the inverse camera matrix $\mathbf{K}^{-1}$. 
This reduces the direct dependence of the performance on the particular camera intrinsics, but this does not make SON camera agnostic as resolution and blur still have an effect. 

\subsubsection{Attention pooling}
Once the descriptors and position encodings have been augmented by the transformers for both images, they are concatenated into a single vector $\mathbf{c} \in \mathbb{R}^{N_m \times 2 \cdot (256+128)}$. 
Because $N_m$ varies between image pairs, a fixed width MLP requires a permutation-invariant pooling operation. 
So, a multi-token attention pooling layer is added which reduces the matches dimension from $N_m$ to a chosen $N_p$. 
A learned attention pooling operation is applied along the matches direction,
\begin{equation}
    \alpha_{pi} = \frac{\exp(\mathbf{q}_p^\top \mathbf{c}_i)}{\sum_{j=1}^{N_m} \exp(\mathbf{q}_p^\top \mathbf{c}_j)}, \, i = 1 \dots N_m
\end{equation}
where the query direction $\mathbf{q}_p$ is learned for $p \in \{1, \dots, N_p\}$ and $\boldsymbol{\alpha}_p = \begin{bmatrix} \alpha_{p1} & \cdots & \alpha_{pN_m} \end{bmatrix}^{\top} \in\mathbb R^{N_m}$ is the attention weight for that particular direction. 
The output feature vector $\mathbf{g}_p$ is taken as the weighted sum along the matches direction, 
\begin{equation}
    \mathbf{g}_p = \sum_{i=1}^{N_m} \alpha_{pi} \mathbf{c}_i \in \mathbb{R}^{2 \cdot (256+128)}
\end{equation}
which is constant in size regardless of the number of feature matches $N_m$.
To get the final output feature vector simply concatenate over $N_p$,
\begin{equation}
    \mathbf{g} = \begin{bmatrix}
        \mathbf{g}_1^\top  & \dots & \mathbf{g}_{N_p}^\top 
    \end{bmatrix}^\top \in \mathbb{R}^{ 2N_p \cdot (256+128)}
\end{equation}
This allows the network to maintain only the most important details of the matched features, and allows for use of the MLP. 

\subsubsection{Pose regression through MLP}
The approach followed by Rockwell et al. \cite{rockwell_8-point_2022} is then used to regress the odometry through an MLP. 
This MLP has two independent heads, each with hidden layers of dimensions $512$ and $128$ and ReLU activations, regressing translation and rotation independently. 
The translation head produces a vector $\mathbf{\hat{u}} \in \mathbb{R}^3$ that is explicitly normalized to unit length.
The rotation head produces a continuous 6D representation of rotation since that provides a better loss landscape for gradient descent \cite{zhou2019continuity}, and is then mapped to $SO(3)$ in the output. 

\subsubsection{Training}
To train SON a pose-loss is constructed which splits the rotation $\mathcal{L}_{rot}$ and translation direction $\mathcal{L}_{tr}$ loss terms,
\begin{equation}
    \mathcal{L} = \frac{\sum_{i} w_b (\Delta \theta_i) \left[ \mathcal{L}_{rot,i} + 4 \mathcal{L}_{tr,i} \right]}{\sum_i w_b (\Delta \theta_i)}  \label{eqn:full_loss}
\end{equation}
where $\mathcal{L}$ is the total loss. 
The bin weight $w_{b(\Delta \theta)}$ is inversely proportional to the frequency of rotation-change magnitude bin $b(\Delta \theta)$, which keeps the most frequent bins from dominating training. 
The rotation and translation direction losses are calculated based on the reference truth rotation $\mathbf{R}$ and translation direction $\mathbf{\hat{u}}$,
\begin{equation}
    \mathcal{L}_{rot,i} = \frac{\left\| \mathbf{\tilde{R}}_i - \mathbf{R}_i \right\|_F^2}{\operatorname{max}(\Delta \theta_i, 2^\circ)}, \quad \mathcal{L}_{tr} = 1 - \mathbf{\tilde{\hat{u}}}^\top \mathbf{\hat{u}} \label{eqn:son_component_losses}
\end{equation}
with the estimates from the network denoted via the tilde superscript. 
The division by $\operatorname{max}(\Delta \theta, 2^\circ)$ emphasizes relative accuracy for small-angle pairs, with a $2^\circ$ floor bounding this contribution. 
This then keeps large rotation pairs, which have larger absolute errors, from dominating the training.

%%%%%%%%%%%%%%%%%%%%%%%%%%%%%%%%%%%%%%
\section{Range observability} \label{sec:scale_observability}
%%%%%%%%%%%%%%%%%%%%%%%%%%%%%%%%%%%%%%
Although the features extracted by SuperPoint, then matched by LightGlue, and finally regressed into an odometry by SON are necessary in order to estimate the orientation and motion of the target, they provide no insight into how far the target is. 
Unlike on Earth, where terrestrial users can rely on objects being ``human"-sized or at least providing scale diversity in busy city streets, satellites provide very little scale diversity (they are the only thing in the image) and are not homogeneous in size (they range from centimeters to tens of meters). 
As such, other methods are required to estimate the range which is necessary as it provides direct insight into the distance to the target, thus informing collision avoidance. 
\hr{The nominal pipeline introduced in this work does not estimate the range: the ROE are tracked up to scale (\autoref{sec:msckf}), which still captures the shape of the relative orbit, including the relative eccentricity and inclination vectors that determine passive safety. }

To resolve the range\hr{,} the filter needs a known metric displacement of the camera center transverse to the line of sight.
In current angles-only literature three options are available to achieve this: orbit maneuvers \cite{woffinden2009observability}, attitude maneuvers \cite{geller2014angles}, and the leveraging of the non-linear dynamics \cite{sullivan2016improved}. 
The latter is the only one which does not rely on changing the spacecraft operations, but it is only available at large distances. 
At the distances considered in this work the relative dynamics are near-linear, and so the nonlinearities are too weak to be relied upon. 

\hr{Attitude maneuvers provide range observability only if the camera has an offset with the center of mass, and thus provide the metric displacement as desired. 
Commanding these maneuvers can make nominal spacecraft operations, such as power generation and communications, difficult or impossible.
This work uses orbit maneuvers to resolve the range, as they can be scheduled around nominal operations and are often already included in them as part of the rendezvous.}

\hr{Later works will investigate what kinds of relative orbits and reconfigurations lead to the most observability, but for this work a single example is used. 
The servicer begins at a standoff distance with pure along-track separation, and then inserts into a passively safe relative orbit (parallel e/i-vector separation) with a lower standoff over two orbits. 
It then keeps the formation with one burn window every quarter orbit, with the burn schedule pre-determined for station-keeping and for observability. 
This schedule is not claimed to be optimal, and instead serves as a sample to demonstrate that once the maneuvers are added, the range becomes observable under this pipeline.}

%%%%%%%%%%%%%%%%%%%%%%%%%%%%%%%%%%%%%%
\section{Multi-State Constraint Kalman Filtering} \label{sec:msckf}
%%%%%%%%%%%%%%%%%%%%%%%%%%%%%%%%%%%%%%
A Multi-State Constraint Kalman Filter (MSCKF) \cite{mourikis2007multi} fuses SON odometry measurements with relative dynamics and multi-frame geometric measurements.
Because neither a target model nor a canonical target body frame is available, the attitude trajectory is observable only up to a single constant rotation. 
Consecutive SON measurements can therefore accumulate attitude drift, which feature tracks bound over the short term and loop closures over the long term.
Because all measurements are relative, the target frame remains arbitrary. 
During evaluation, one constant rotation aligns the estimated and reference truth attitude histories before attitude errors are computed.  

\subsection{State}
The MSCKF tracks an error state which it then  uses to update the corresponding nominal state. 
The nominal state is made up of the osculating quasi non-singular Relative Orbit Elements (ROE) \cite{koenig2017new} $\delta \boldsymbol{\alpha}$ \hr{projected onto a sphere of fixed radius}, the attitude of the target's body relative to the servicer camera $\mathbf{q}_{T/C}$ represented using a quaternion, the associated angular velocity $^T\boldsymbol{\omega}_{T/C}$, and the norm of the ROE, $s$, which is treated as a consider parameter.
The nominal state vector $\mathbf{x}$ is therefore given as, 
\begin{equation}
    \mathbf{x}^t = \begin{bmatrix}
        \mathbf{q}_{T/C}^{t} & ^T\boldsymbol{\omega}_{T/C}^{t} & a\delta \boldsymbol{\alpha}^t & s
    \end{bmatrix}^\top \label{eqn:reference_state}
\end{equation}
The estimated nominal state on the other hand is represented as, 
\begin{equation}
    \boldsymbol{\mu}^{t\mid t} = \begin{bmatrix}
        \hat{\mathbf{q}}_{T/C}^{t} & ^T\hat{\boldsymbol{\omega}}_{T/C}^{t} & a\hat{\delta \boldsymbol{\alpha}}^t & \hat{s}
    \end{bmatrix}^\top 
\end{equation}
where $\hat{\cdot}$ represents an estimate and $\boldsymbol{\mu}^{t \mid t}$ the filter's mean at time $t$ after the measurement update. 
From the servicer $S$ and target $T$ Keplerian elements $\boldsymbol{\alpha} = \begin{bmatrix} a & e & i & \Omega & \omega & M \end{bmatrix}$, the quasi non-singular ROE are
\begin{equation}
    \delta \boldsymbol{\alpha} = \begin{bmatrix}
        \delta a & \delta \lambda & \delta e_x & \delta e_y & \delta i_x & \delta i_y
    \end{bmatrix}^\top % = \begin{bmatrix}
    %     (a_T - a_S) / a_S \\ (u_T - u_S) + (\Omega_T - \Omega_S) \cos i_S \\ e_T \cos \omega_T - e_S \cos \omega_S \\ e_T \sin \omega_T - e_S \sin \omega_S \\ i_{T} - i_{S} \\ (\Omega_T - \Omega_S) \sin i_S
    % \end{bmatrix}
\end{equation}
where $\delta a$ is the relative semi-major axis, $\delta \lambda$ the relative mean longitude, and ($\delta e_x$, $\delta e_y$) and ($\delta i_x$, $\delta i_y$) the relative eccentricity and inclination vectors, scaled to meters by the servicer semi-major axis a.

\hr{Following work from bearings-only tracking \cite{aidala1983utilization} and monocular SLAM \cite{civera2007dimensionless}, the ROE state is split into its shape and its size.
The shape is estimated as a point on a sphere of radius $s_0 = \| a \hat{\delta \boldsymbol{\alpha}}_0 \| $, which since the size is not estimated is constant and determined by the initial ROE estimate. 
The ROE shape error is then the 5D coordinate in the tangent space of this sphere, expressed in an orthonormal basis $\mathbf{B}^t \in \mathbb{R}^{6 \times 5}$. 
After every propagation and update the estimate is renormalized onto the sphere.
Since monocular measurements cannot distinguish the relative orbit from its reflection behind the camera, the estimate is also reflected whenever the predicted target position falls behind the camera.}

\hr{If the measurements are not a function of $s$, this parameter can be set without taking into account its effects on the covariance. 
If there is a non-zero camera lever arm ${}^S \mathbf{r}_{C/S}$ and any attitude motion, then the lever arm's size relative to the relative orbit does depend on $s$.
To maintain generality, $s$ is not estimated, but it is included as a Schmidt consider parameter \cite{schmidt1966application}. 
Its uncertainty is carried into the covariance and enters the measurement Jacobians, but its estimate is never updated such that $\hat{s} = s_0$.
Its prior variance is the initial ROE variance along the estimated direction $\sigma_s^2 = \hat{\mathbf{d}}_0^\top \Sigma_{\delta \alpha, 0} \hat{\mathbf{d}}_0$ with $\hat{\mathbf{d}}_0 = a \hat{\delta \boldsymbol{\alpha}}_0 / s_0$. 
Note that since this is a consider parameter, in every update the row of the Kalman gain corresponding to $s$ is set to zero.}

The error state is then a delta of the nominal state, where the quaternion is mapped to a rotation vector (as is done in the original MSCKF \cite{mourikis2007multi}) and the \hr{angular velocity and consider parameter are simple differences}, 
\begin{equation}
    \Delta \mathbf{x}^t = \begin{bmatrix}
        \Delta \boldsymbol{\theta}_{T/C}^{t} & \Delta {}^T \boldsymbol{\omega}_{T/C}^{t} & \Delta a\delta \boldsymbol{\alpha}^t & \Delta s
    \end{bmatrix}^\top
\end{equation}
The nominal attitude from \autoref{eqn:reference_state} is related to the rotation vector via, 
\begin{equation}
    \mathbf{q}_{T/C}^{t} = \hat{\mathbf{q}}_{T/C}^{t} \otimes \delta q \left(\Delta \boldsymbol{\theta}_{T/C}^{t} \right) \label{eqn:quaternion_to_rotation_vector}
\end{equation}
where $\delta q (\cdot)$ converts the input rotation vector to the error quaternion as,
\begin{equation}
    \delta q (\Delta \theta) = \begin{bmatrix}
        \cos \left( \frac{\| \Delta\theta\|}{2}\right) & \sin \left( \frac{\|\Delta\theta\|}{2}\right)  \frac{\Delta \theta^\top}{\| \Delta \theta \|}
    \end{bmatrix}^\top
\end{equation}

In addition to utilizing an error state, the MSCKF uses attitude clones so that measurements relating past and current images can be expressed as functions of the current augmented state \cite{mourikis2007multi}. 
This is needed for both the feature tracks and the loop closures, so two sets of attitude clones are maintained. 
The first set contains the $N_w$ most recent non-keyframe attitude clones retained by the sliding window,
\begin{equation}
    \mathcal{W}^t = \begin{Bmatrix}
        \tau_1 & \tau_2 & \dots & \tau_{N_w}
    \end{Bmatrix}
\end{equation}
and the second set includes a set of $N_k$ keyframes retained to perform loop closure, 
\begin{equation}
    \mathcal{K}^t = \begin{Bmatrix}
        \kappa_1 & \kappa_2 & \dots & \kappa_{N_k}
    \end{Bmatrix}
\end{equation}
These sets do not overlap, and as such $\mathcal{W}^t \cap \mathcal{K}^t = \varnothing$. 
With this, the augmented nominal and error states are,
\begin{equation}
    \mathbf{x}^t_{\text{aug}} = \begin{bmatrix}
        \mathbf{q}_{T/C}^{t} \\ ^T\boldsymbol{\omega}_{T/C}^{t} \\ a\delta \boldsymbol{\alpha}^t \\ s \\ \mathbf{q}_{T/C}^{\tau}, \tau \in \mathcal{W}^t \\ \mathbf{q}_{T/C}^{\tau}, \tau \in \mathcal{K}^t
    \end{bmatrix}
\end{equation}
\begin{equation}
    \Delta \mathbf{x}^t_{\text{aug}} = \begin{bmatrix}
        \Delta \boldsymbol{\theta}_{T/C}^{t} \\ \Delta {}^T \boldsymbol{\omega}_{T/C}^{t} \\ \Delta a\delta \boldsymbol{\alpha}^t \\ \Delta s \\ \Delta \boldsymbol{\theta}_{T/C}^{\tau}, \tau \in \mathcal{W}^t \\ \Delta \boldsymbol{\theta}_{T/C}^{\tau}, \tau \in \mathcal{K}^t
    \end{bmatrix}
\end{equation}
where the newest clone is initialized as a copy of the current attitude state.
$\mathcal{W}^t$ and $\mathcal{K}^t$ are the sliding window and selected keyframes from \autoref{fig:overall_pipeline}.
At initialization, \(\mathcal{W}^0\) contains the initial attitude clone and \(\mathcal{K}^0=\varnothing\), both sets subsequently grow up to their configured limits.
In general these are abbreviated by $N_c = N_w + N_k$ which determines the number of attitude clones. 

Only the attitude is cloned, and when necessary for the measurement function the ROE at previous time steps are back-propagated from the current time estimate using the inverse to the STM used for dynamics propagation, discussed in the following section. 

The covariance is also augmented via a selection matrix $\mathbf{J}$,
\begin{equation}
    \mathbf{J} = \begin{bmatrix}
        \mathbf{I}_{3} & \mathbf{0}_{3 \times (12+3N_c-3)}
    \end{bmatrix}
\end{equation}
where $12 + 3N_c$ is the size of the augmented state before the new clone is appended. 
This selection matrix singles out the attitude error from the augmented error state, and after cloning the covariance is updated as, 
\begin{equation}
    \boldsymbol{\Sigma}^{t \mid t}_{\text{aug}} = \begin{bmatrix}
        \boldsymbol{\Sigma}^{t \mid t} & \boldsymbol{\Sigma}^{t \mid t} \mathbf{J} ^\top \\ \mathbf{J} \boldsymbol{\Sigma}^{t \mid t} & \mathbf{J} \boldsymbol{\Sigma}^{t \mid t} \mathbf{J}^\top
    \end{bmatrix}
\end{equation}
This operation holds for all steps of the MSCKF, where $N_c$ begins at $1$ for the first step and then grows to the maximum set value after sufficient attitude states have been estimated. 

\subsection{Dynamics model}
As previously mentioned, the dynamics are applied directly to the nominal state, and the error state is obtained from this dynamically propagated state $\boldsymbol{\mu}^{t \mid t-1}$. 
To propagate the dynamics of the nominal state over the propagation interval $\Delta t$ it must be split into its principal components: attitude, angular velocity, and ROE. 
ROE are propagated in the filter using a J2-only state transition matrix $\boldsymbol{\Phi}_{NS}$ which is appropriate for small spacecraft separations \cite{koenig2017new},
\begin{equation}
    a \hat{\delta \boldsymbol{\alpha}}^{t \mid t-1} = \boldsymbol{\Phi}_{NS}^{t}(\boldsymbol{\alpha}_{S}^{t}, \Delta t) a \hat{\delta \boldsymbol{\alpha}}^{t-1\mid t-1}
\end{equation}

The attitude is propagated kinematically, 
\begin{equation}
    \hat{\mathbf{q}}_{T/C}^{t \mid t-1} = \hat{\mathbf{q}}_{T/C}^{t-1 \mid t-1} \otimes \delta \mathbf{q}(^T\hat{\boldsymbol{\omega}}_{T/C}^{t-1 \mid t-1} \Delta t)
\end{equation}
\begin{equation}
    ^T\hat{\boldsymbol{\omega}}_{T/C}^{t \mid t-1} = {}^T\hat{\boldsymbol{\omega}}_{T/C}^{t-1 \mid t-1} \label{eqn:kinematic_ang_vel}
\end{equation}
This design choice is made to avoid estimating the inertia components and the associated principal axis, which is a task that is beyond the scope of this work. 
The inertia is therefore neither estimated nor assumed known.

To arrive at the first-order Jacobian for propagating the covariance, first model the attitude increment over $\Delta t$ as,
\begin{equation}
    \delta \mathbf{q}_{\omega}^{t \mid t-1} = \delta \mathbf{q}({}^T\hat{\boldsymbol{\omega}}_{T/C}^{{t-1\mid t-1}} \Delta t)
\end{equation}
such that,
\begin{equation}
    \boldsymbol{\Phi}^{t \mid t-1} = \begin{bmatrix}
        R(\delta \mathbf{q}_{\omega}^{t\mid t-1})^\top & \Delta t \mathbf{I}_{3} & 0 & 0 \\
        0 & \mathbf{I}_{3} & 0 & 0 \\
        0 & 0 & \tilde{\boldsymbol{\Phi}}_{NS}^{t} & 0 \\
        0 & 0 & 0 & \mathbf{I}_{3N_c+1}
    \end{bmatrix} \label{eqn:stm_augmented}
\end{equation}
where the ROE block is \hr{$\tilde{\boldsymbol{\Phi}}_{NS}^{t} = (\boldsymbol{B}^t)^\top \boldsymbol{\Phi}_{NS}^{t} \boldsymbol{B}^{t-1} / g$ is the J2 STM projected onto the tangent space $\mathbf{B}$, with $g$ the norm growth over the step. }
The consider parameter and the attitude clones are constant, so their propagation is identity.
The covariance is updated as,
\begin{equation}
    \bar{\boldsymbol{\Sigma}}^{t \mid t-1} = \boldsymbol{\Phi}^{t \mid t-1} \boldsymbol{\Sigma}^{t-1 \mid t-1} (\boldsymbol{\Phi}^{t \mid t-1})^\top + \mathbf{Q}_{\text{aug}} \label{eqn:ekf_covariance_propagation}
\end{equation}
where the augmented process noise is,
\begin{equation}
    \mathbf{Q}_{\text{aug}} = \text{diag}( \begin{bmatrix}
        Q_\theta \mathbf{I}_{3} & Q_\omega \mathbf{I}_{3} & (B^t)^\top Q_{ROE} \mathbf{I}_{6} B^t & \mathbf{0}_{3N_c + 1 \times 3N_c + 1}
    \end{bmatrix})
\end{equation}
Previously cloned attitude states \hr{and the consider parameter are} static and hence do not receive additional process noise by design. 

\subsection{SON measurement model}
In the MSCKF update step, when a SON measurement $\mathbf{y}_{\text{SON}}^t = (\mathbf{y}^t_q, \mathbf{y}^t_{\hat{u}})$ is available, the first step is to use the state estimates to produce a measurement estimate $\hat{\mathbf{y}}_{\text{SON}}^t$. 
As in \autoref{eqn:odometry_quaternion_update}, the previous $\hat{\mathbf{q}}_{T/C}^{t-1}$ and current state estimate $\hat{\mathbf{q}}_{T/C}^{t}$ are used to compute the modeled measurement $\hat{\mathbf{y}}^{t \mid t-1}_q$,
\begin{equation}
    \hat{\mathbf{y}}^{t \mid t-1}_q = \hat{\mathbf{q}}_{T/C}^{t} \otimes (\hat{\mathbf{q}}_{T/C}^{t-1})^* 
\end{equation}
The residual $\boldsymbol{\nu}^{t}_q$ is computed from the SON measurement $\mathbf{y}^t_q$,
\begin{equation}
    \boldsymbol{\nu}^{t}_q = \log \left( \left( \hat{\mathbf{y}}^{t \mid t-1}_q \right)^* \otimes \mathbf{y}^t_q \right)
\end{equation}
and is linearized numerically as,
\begin{equation}
    \boldsymbol{\nu}^{t}_q \approx \mathbf{H}_{q}^{t} \Delta \mathbf{x}^{t}_{\text{aug}}
\end{equation}

Before accounting for the translation direction measurement from SON, the filter performs the measurement update step on the rotation measurement $\mathbf{y}^t_q$.
Here the previously calculated measurement Jacobian along with the Joseph form of the Kalman filter \cite{bucy2005filtering}, are used to update the mean state $\boldsymbol{\mu}^{t \mid t}$ and covariance $\boldsymbol{\Sigma}^{t \mid t}$. 

The translation direction must now be modeled from the estimated state.
To this end, the ROE are converted to a Radial Tangential Normal (RTN) position using a linear mapping $F$. 
This linear mapping is a non-linear function of the absolute orbit elements of the servicer \cite{sullivan2017nonlinear}, so it is denoted as $F(\boldsymbol{\alpha}_{S}^{t})$.
The RTN position of the target $T$ with respect to the servicer $S$ is then,
\begin{align}
    ^S\hat{\mathbf{r}}_{T/S}^{t \mid t-1} = F(\boldsymbol{\alpha}_{S}^{t}) \kappa^t a \hat{\delta \boldsymbol{\alpha}}^{t \mid t-1} \label{eqn:roe_to_rtn}
\end{align}
\hr{where $\kappa^t = \hat{s} / \| \boldsymbol{\Psi}^{t_0,t} a\hat{\delta \boldsymbol{\alpha}}^{t} \| $ rescales the estimate to the ROE norm.
Since the ROE dynamics are linear and the estimate is only renormalized, the direction is back-propagated to $t_0$, where its norm is $\hat s$.}
This relative position is then expressed in the camera frame $C$ through the following linear transformation,
\begin{align}
    ^C\hat{\mathbf{r}}_{T/C}^{t \mid t-1} = \mathbf{R}_{C/S}^{t} ({}^S\hat{\mathbf{r}}_{T/S}^{t \mid t-1} - {}^S\mathbf{r}_{C/S}^{t}) \label{eqn:rtn_to_camera_frame}
\end{align}
where $\mathbf{R}_{C/S}^{t}$ and ${}^{S}\mathbf{r}_{C/S}^{t}$ are the camera attitude and lever arm expressed in the RTN frame at time $t$.
Since the translation-direction update is applied after the rotation update, the quantities below are evaluated at the state after that update, denoted by the superscript $+$.
Previous ROE are not cloned, so the ROE at the time of the previous images are back-propagated from this estimate,
\begin{equation}
    a \hat{\delta \boldsymbol{\alpha}}^{t-1, +} = (\boldsymbol{\Phi}_{NS}^{t})^{-1} a \hat{\delta \boldsymbol{\alpha}}^{t,+}
\end{equation}
Applying the same mappings yields ${}^{C}\hat{\mathbf{r}}_{T/C}^{t-1,+}$.
The resulting translation change expressed in the current camera frame is, 
\begin{equation}
    \hat{\Delta \mathbf{r}}^{t,+} = {}^C\hat{\mathbf{r}}_{T/C}^{t,+} - \hat{\mathbf{R}}^+ {}^C\hat{\mathbf{r}}_{T/C}^{t-1,+}
\end{equation}
where $\hat{\mathbf{R}}^+$ is the rotation matrix representation of $\hat{\mathbf{y}}^{+}_q$ which is now re-calculated based on the measurement update. 
The normalized unit direction $\hat{\mathbf{u}}^{t,+}$ is then,
\begin{equation}
    \hat{\mathbf{y}}^{t,+}_{\hat{u}} = \frac{\hat{\Delta \mathbf{r}}^{t,+}}{\| \hat{\Delta \mathbf{r}}^{t,+} \|}
\end{equation}

The residual for the translation unit direction $\boldsymbol{\nu}_{\hat{u}}^t$ is calculated from the measured translation direction from SON $\mathbf y^t_{\hat{u}}$,
\begin{equation}
    \boldsymbol{\nu}_{\hat{u}}^t = \mathbf{y}^t_{\hat{u}} - \hat{\mathbf{y}}^{t,+}_{\hat{u}} 
\end{equation}
which is again linearized numerically as,
\begin{equation}
    \boldsymbol{\nu}_{\hat{u}}^t \approx \mathbf{H}_{\hat{u}}^{t} \Delta \mathbf{x}^{t}_{\text{aug}}
\end{equation}
\hr{The direction of a small translation is poorly measured, so the $\hat{\mathbf{u}}$ covariance grows with the inverse of the predicted parallax $\theta = \| \hat{\Delta \mathbf{r}}^{t,+} \| / \| {}^C\hat{\mathbf{r}}_{T/C}^{t,+} \|$, as $\sigma_{\hat{u}}^2 = \sigma_{\hat{u},0}^2 + (k_{\hat{u}}/\theta)^2$. The constants $\sigma_{\hat{u},0}$ and $k_{\hat{u}}$ (\autoref{tab:simulation_parameters}) are tuned on SON errors on Monte Carlo trajectories.}

The measurement update is again completed via the calculated Jacobian for the translation direction and the Joseph form to arrive at the current state estimate. 

\subsection{Feature tracks measurement model}

A feature observation is an image location $\mathbf{y}_\ell^t = \begin{bmatrix}u_\ell^t & v_\ell^t  \end{bmatrix}^\top$ associated with a distinctive point on the target surface, whereas a feature track is the sequence of image coordinates of length $c$, $\mathbf{y}_\ell = \begin{bmatrix} \mathbf{y}_\ell^\tau \end{bmatrix}, \tau \in \{ t, t-1, \dots, t - c +1\}$, believed to correspond to the same physical target point in several images. 
Tracks are thus constructed by running SuperPoint on the current image, and matching features with the previous one.
If a feature has been matched in at least $c_{\text{min}}$ images, it is considered a track, and remains as such until it goes out of view or it is about to leave the clone window $N_w$, such that $c \leq N_w$. 
A minimum track length is set because each track observed in $c$ images provides $2c-3$ constraints ($2c$ residuals and 3 unknowns from the feature position), so if $c$ is small the feature track triangulates poorly.
For each completed track, a nominal feature position $^{T}\hat{\mathbf r}_{\ell/T}$ is first triangulated using the cloned attitudes and the ROE-derived camera positions. 
This position is used only to form the residual and its Jacobians, and is never added to the filter state.
The attitude clone and feature position blocks follow the original MSCKF \cite{mourikis2007multi}, but the ROE block, which back-propagates the current ROE to each observation time $\tau$, is
\begin{equation}
    \mathbf{H}_{\delta \boldsymbol{\alpha}, \ell}^{\tau} = \kappa^t \mathbf{J}_{\pi}^{\tau} \mathbf{R}_{C/S}^{\tau}  F(\boldsymbol{\alpha}_{S}^{\tau}) \boldsymbol{\Psi}^{\tau, t} \mathbf{B}^t
\end{equation}
where $\mathbf{J}_{\pi}^{\tau}$ is the pinhole projection Jacobian evaluated at the predicted feature position and  $\boldsymbol{\Psi}^{\tau,t} = (\boldsymbol{\Phi}_{NS}^{\tau,t})^{-1}$ is the inverse of the STM over the time interval from $\tau$ to $t$.
\hr{The consider parameter's column is given as,
\begin{equation}
    \mathbf{H}^\tau_{s,\ell} = \frac{\kappa^t}{\hat s} \mathbf{J}^\tau_\pi \mathbf{R}^\tau_{C/S} F(\boldsymbol{\alpha}^\tau_S) \boldsymbol{\Psi}^{\tau,t} a\hat{\delta\boldsymbol{\alpha}}^t
\end{equation}
where the dependence of $\kappa^t$ on $a \hat{\delta \boldsymbol{\alpha}}^t$ is included in the ROE block.}

As in the original MSCKF \cite{mourikis2007multi}, the feature position is removed by projecting the stacked residual onto an orthonormal basis of the left null space of its feature position Jacobian.
For a rank-three feature position Jacobian, this gives the $(2c-3)$-dimensional linearized residual $\boldsymbol{\nu}_\ell^0$, 
\begin{equation}
    \boldsymbol{\nu}_\ell^0  \approx \mathbf{H}_\ell^0 \Delta \mathbf{x}^t_{\text{aug}} \label{eqn:lin_residual_feature_tracks}
\end{equation}
with measurement covariance $C_\ell^0 = C_f^2 \mathbf{I}_{2c-3}$.

Like with SON, once this residual is available, the related Jacobian $\mathbf{H}_\ell^0$ and the Joseph form are used to update the state based on this single feature track. 

\subsection{Loop closures}
Loop closure is implemented by retaining selected attitude clones $\mathcal{K}^t$ after they leave the normal sliding window $\mathcal{W}^t$. 
After a warm-up period, a clone is promoted to a keyframe if its estimated attitude is at least $15^\circ$ from that of every other retained keyframe until the maximum number of cloned keyframes $N_k$ is reached. 
At that point, a candidate replaces the keyframe closest to its nearest neighbor whenever this increases the minimum angular separation of the set $\mathcal{K}^t$.
These keyframes remain correlated filter states and can be corrected by subsequent measurement updates to ensure that the filter remains self-consistent. 

The loop-closure measurement $\mathbf y_{\mathrm{LC},q}^{t,\kappa}$ is obtained by running the same SuperPoint, LightGlue, and SON pipeline between the retained keyframe and current image.
Only SON's rotation output is used for loop closure, with measurement covariance $\mathbf C_{\mathrm{LC}}$.

To determine whether or not to perform loop closure between the current image and one of the keyframes in $\mathcal{K}^t$, a threshold is used.
A retained keyframe is considered only when it is at least \(2N_w\) frames old and its predicted rotation relative to the current frame is no greater than \hr{\(45^\circ\)}.
The loop-closure update is accepted when at least 100 LightGlue matches are available and its Normalized Innovation Squared (NIS) is no greater than 40.
The innovation gate is deliberately loose, as once drift has accumulated a correct loop closure would produce a large innovation, and so a tighter gate may reject the measurement.
The rotation threshold \hr{keeps loop pairs within the $45^\circ$ viewpoint change of SON's training pairs, beyond which SON under-estimates the rotation, and protects} against runaway compute, since otherwise the odometry estimation portion of the pipeline would run for all $N_k$ keyframes on every image. 
The recency requirement excludes keyframes which were recently promoted to $\mathcal{K}^t$, as these do not add more constraints than the feature tracks.
Useful loop closures then are those that occur when the target's rotation brings a previously observed side back into view after a long interval.

An accepted loop closure removes the drift accumulated since the keyframe was last used, hence allowing for consistency in the attitude history. 

\begin{figure*}[t]
    \includegraphics[width=0.75\textwidth]{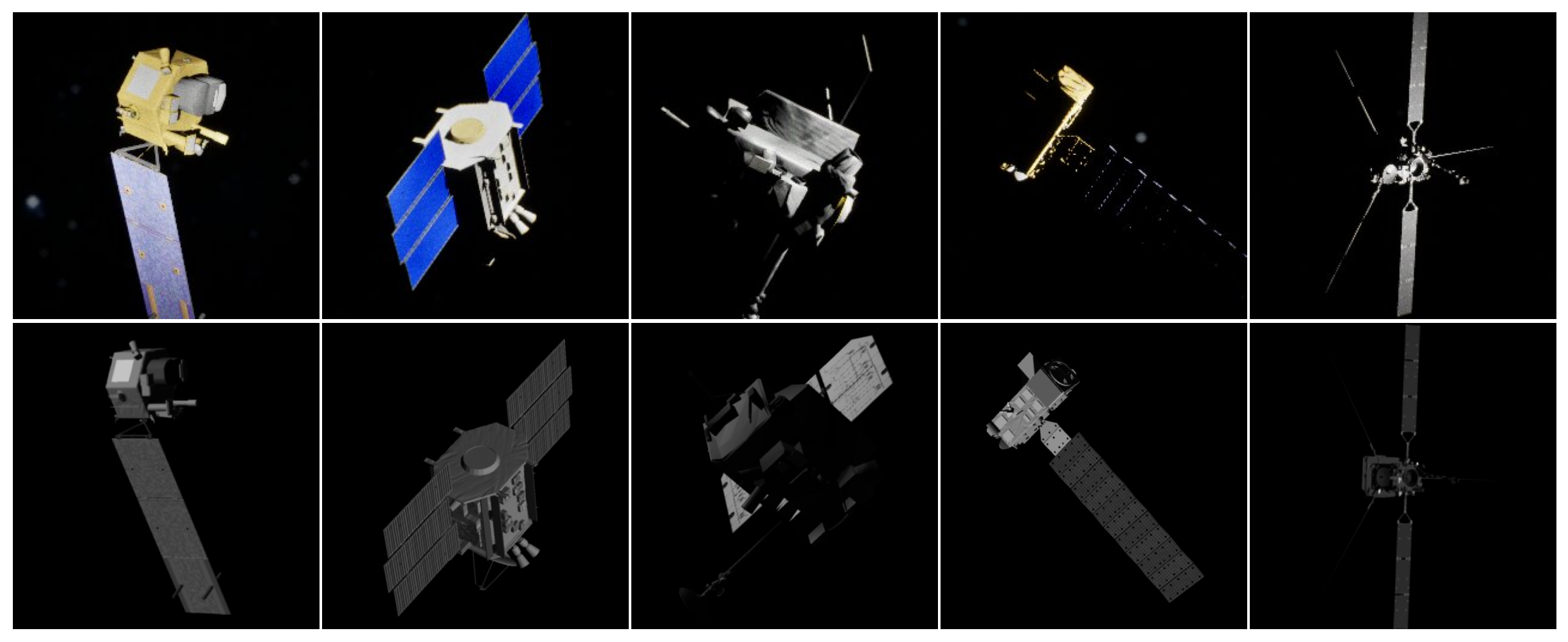}
    \centering
    \caption{Sample comparison images of the original SPE3R dataset (top) and the new re-rendered version SPE3R+ (bottom). Note that the re-rendered versions of the images are cropped in this figure to match the scale of the original dataset. \vspace{-0.5cm}}
    \label{fig:spe3r_comparison}
\end{figure*}

\subsection{ROE fading-memory covariance inflation}

The SON and feature-track errors are temporally and mutually correlated, whereas the filter measurement model treats them as independent. 
To mitigate the resulting contraction of the ROE covariance, a fading-memory covariance inflation is applied after each conventional covariance propagation in \autoref{eqn:ekf_covariance_propagation}. 
Define the inflation matrix as a block diagonal matrix,
\begin{equation}
    \mathbf{D}_{\delta \boldsymbol{\alpha}} = \operatorname{blkdiag} \left( \begin{bmatrix}
        \mathbf{I}_{6} & \sqrt{\lambda_{\rho}}\mathbf{I}_{5} & \mathbf{I}_{1+3N_c}
    \end{bmatrix} \right)
\end{equation}
where the first six states are the attitude and angular-velocity errors, the subsequent $5$ are the ROE errors, \hr{followed by the consider parameter}, and $N_c$ is the total number of active attitude clones, including retained loop-closure keyframes. 
The inflation matrix $\mathbf{D}_{\delta \boldsymbol{\alpha}}$ leaves the attitude, \hr{consider,} and clone blocks untouched, scaling the ROE block by $\lambda_\rho$ and the cross-covariances by $\sqrt{\lambda_\rho}$.
The covariance used for the subsequent measurement updates is
\begin{equation}
    \boldsymbol{\Sigma}^{t|t-1} = \mathbf{D}_{\delta \boldsymbol{\alpha}} \bar{\boldsymbol{\Sigma}}^{t|t-1} \mathbf{D}_{\delta \boldsymbol{\alpha}}^{\top}
\end{equation}

\subsection{Measurement gating}
The SON rotation and translation-direction estimates are processed sequentially with a bounded-influence update gate.
For a SON innovation $\nu$ with measurement covariance $C$, projected innovation covariance $S = H\Sigma^{t|t-1}H^\top + C$, and $\gamma = \nu^\top S^{-1}\nu$, the update uses
\begin{equation}
  \tilde{C} = \max\!\left(1,\ \gamma/\bar{\gamma}\right) C, \qquad \bar{\gamma} = \chi^2_3(0.99),
\end{equation}
so that an outlier's influence is bounded without the measurement being discarded.
If fewer than eight LightGlue matches are available, SON does not produce a measurement and the corresponding channel is skipped. 

For a feature track containing $c$ accepted observations, the projected innovation covariance and normalized innovation statistics are
\begin{align}
    \mathbf{S}^{0}_{\ell} &= \mathbf{H}^{0}_{\ell} \boldsymbol{\Sigma}^{t|t-1} \left( \mathbf{H}^{0}_{\ell} \right)^{\top} + \mathbf{C}^{0}_{\ell} \\
    \gamma^{0}_{\ell} &= \left( \boldsymbol{\nu}^{0}_{\ell} \right)^{\top} \left( \mathbf{S}^{0}_{\ell} \right)^{-1}  \boldsymbol{\nu}^{0}_{\ell}
\end{align}
The track is accepted when
\begin{equation}
    \gamma^{0}_{\ell} \leq 0.5 \chi_{2c-3}^{2}(0.95).
\end{equation}
This tight gate was selected via trial and error through Monte Carlo simulations.

%%%%%%%%%%%%%%%%%%%%%%%%%%%%%%%%%%%%%%
\section{Training \& Implementation Details} \label{sec:training_and_implementation_details}
%%%%%%%%%%%%%%%%%%%%%%%%%%%%%%%%%%%%%%

\subsection{Key assumptions}
Five key assumptions are made that limit the immediate applicability of this work in flight missions. 
\begin{enumerate}
    \item \textbf{Sensing}: The camera boresight remains aligned with the target, and an upstream detector supplies a target bounding box. The present evaluation derives this bounding box from a dilated reference truth mask. Future work will replace the reference truth cropping and pointing with a detector and estimator-based control.
    \item \textbf{Servicer knowledge}: The servicer position and attitude in the Earth-centered inertial (ECI) frame are known exactly. \hr{Camera intrinsics and position on the bus are exact and do not change during simulation. Target pointing and the known maneuvers of \autoref{sec:scale_observability} are executed without error or delay.} In flight, these quantities are supplied by GNSS, star trackers, and calibration, typically orders of magnitude more accurately than the relative estimates sought here. An upstream method is assumed to initialize the ROE state of the filter. 
    \item \textbf{Image domain}: Images are synthetic and contain neither sensor noise nor an Earth background. A sun-synchronous orbit is simulated, with the target illuminated throughout the trajectory. The odometry front-end is individually evaluated on hardware-in-the-loop images captured under space-like illumination in the TRON facility, while the complete filter is evaluated on rendered images only. 
    No formal calibration against flight imagery is available.
    \item \textbf{Target}: The target is assumed to be rigid, without articulating appendages. Sufficient repeatable surface features remain visible across the baselines used by the measurements.
    \item \textbf{Compute}: Sufficient compute is assumed to be available. Inference benchmarking is performed on an NVIDIA AGX Orin which is an edge device, but does not necessarily represent current flight-ready hardware.
\end{enumerate}

\subsection{Datasets}

Four evaluation datasets are used in this work.
Two of these are used just in the evaluation of LightGlue as is done in the original work by Lindenberger et al. \cite{lindenberger_lightglue_2023}, and so only include terrestrial scenes.
In contrast, the other datasets (SPE3R+, SHIRT) include images of satellites only. 
In addition to these static datasets, trajectories are generated and rendered on the fly for testing the MSCKF with SON in the loop.

\subsubsection{HPatches}
HPatches \cite{balntas2017hpatches} is a dataset commonly used in literature to evaluate feature matchers which includes 5 image pairs per sequence with known homographies.
The precision (P) when compared against the reference truth matches at a reprojection error of 3 px is reported.
The accuracy of the estimated homographies from the feature matches using both RANSAC \cite{fischler1981random} and weighted Direct Linear Transformation (DLT) \cite{hartley_multiple_2004} is also reported. 
The accuracy is reported by computing the mean reprojection error of the four image corners and reporting the area under the cumulative error curve (AUC) up to values of 1 px and 5 px. 

\subsubsection{MegaDepth1500}
MegaDepth \cite{li2018megadepth} provides landmark imagery with reconstructed camera calibrations and poses. 
Evaluation uses the MegaDepth1500 set which includes 1,500 image pairs from two of the scenes to assess matching and relative-pose accuracy. 
As in the baseline described in \autoref{sec:odometry_estimation_from_feature_matches}, the feature matches are used to compute the essential matrix within a RANSAC framework \cite{fischler1981random}. 
The pose error is computed as the maximum angular error in rotation and translation and then the AUC is reported at different angular separations. 
Together HPatches and MegaDepth1500 show how much terrestrial performance fine-tuning costs. 

\subsubsection{SPE3R+}

\begin{table*}[t]
\centering
\caption{LightGlue and Spacecraft Odometry Network (SON) training configuration.}
\footnotesize
\setlength\tabcolsep{4.0pt}%
\renewcommand{\arraystretch}{1.16}
\setlength{\heavyrulewidth}{0.4pt}
\setlength{\lightrulewidth}{0.4pt}
\setlength{\cmidrulewidth}{0.4pt}
\begin{tabular}{r r l l}
\toprule
Group & Parameter & LightGlue & SON \\
\midrule
\multirow{3}{*}{Data}
& Crop
& Margin 1.3; $768\times768$ px
& Same crop; cached matches \\
& Input
& \makecell[l]{Frozen SuperPoint features;\\$\dim(\mathbf d_i)=256$; $\leq2048$ features}
& \makecell[l]{LightGlue matches;\\$\mathbf d_i$, $\mathbf p_i$} \\
& Pairs
& \makecell[l]{120,000/epoch; total rotation change $\leq120^\circ$;\\bins $\{0,2,4,6,8,12,20,30, 45, 60, 75, 90, 120\}^\circ$}
& \makecell[l]{393,946 train pairs; viewing direction change$\leq45^\circ$;\\ $\{0, 2, 4, 6, 8, 12, 20, 30, 45\}^\circ$  rotation bins} \\
\midrule
\multirow{2}{*}{Model}
& Backbone
& Standard LightGlue
& $N_t=9$; 4 heads; $N_p=4$ \\
& Head
& Match-assignment head
& Two MLP heads; widths $512$, $128$ \\
\midrule
\multirow{1}{*}{Loss}
& Objective
& \makecell[l]{Metric depth and pose;\\3/5 px match thresholds; 5 px epipolar}
& Equations \ref{eqn:full_loss}--\ref{eqn:son_component_losses} \\
\midrule
\multirow{3}{*}{Opt.}
& Optimizer
& Adam \cite{kingma2014adam}
& AdamW \cite{loshchilov2017decoupled} \\
& Schedule
& \makecell[l]{50 epochs; $\eta_0=10^{-4}$;\\exponential decay after epoch 30}
& \makecell[l]{30 epochs; $\eta_0=2\times10^{-5}$;\\2-epoch warm-up; cosine to $10^{-7}$} \\
& Batch/reg.
& $B=8$, no decay or clipping
& $B=32$; weight decay $10^{-4}$; gradient clip 2 \\
\bottomrule
\end{tabular}

\label{tab:training_parameters}
\end{table*}

\begin{table*}[t]
\centering
\caption{Simulation, control, and filter configuration. Noise entries are per-component process variances or measurement standard deviations, as defined in the text. }
\footnotesize
\setlength\tabcolsep{4.0pt}%
\renewcommand{\arraystretch}{1.14}
\setlength{\heavyrulewidth}{0.4pt}
\setlength{\lightrulewidth}{0.4pt}
\setlength{\cmidrulewidth}{0.4pt}
\begin{tabular}{r r l l}
\toprule
Group & Quantity & Symbol & Value \\
\midrule
\multicolumn{4}{c}{\textit{Truth, imaging, and control}} \\
\cmidrule(lr){1-4}
\multirow{2}{*}{Orbit}
& Initial elements & $\boldsymbol\alpha_{S,0}$
& \makecell[l]{$a=7078.135$ km, $e=0.001$, $i=98.19^\circ$,\\$\Omega=189.89086^\circ$, $\omega=M=0^\circ$} \\
& Epoch & $t_0$ & 2011-07-18 01:00:00 \\
Truth
& Dynamics & --
& \makecell[l]{$120 \times 120$ gravity field; drag; SRP; Sun/Moon gravity;\\gravity-gradient, drag, and SRP torques} \\
\multirow{2}{*}{Image}
& Cadence & $\Delta t$ & $5$ s \\
& Camera lever arm & ${}^S \textbf{r}_{C/S}$ & $[0.10, 0.15, 0.41]$ m \\
\midrule
\multicolumn{4}{c}{\textit{MSCKF}} \\
\cmidrule(lr){1-4}
\multirow{1}{*}{Dynamics}
& Propagation & $(\boldsymbol\Phi_{NS},\mathbf q)$ & $J_2$ ROE STM; kinematic attitude \\
Process
& Noise & $(Q_\theta,Q_\omega,Q_{\mathrm{ROE}})$ & $(10^{-4} \text{rad}^2,10^{-5} \text{rad}^2\text{/s}^2,[10^{-6}, 10^{-5}, 10^{-6}, 10^{-6}, 10^{-6}, 10^{-6}] \text{m}^2)$ \\
\multirow{2}{*}{Measurement}
& SON & $(C_q,C_{\hat u})$ & \hr{$(1.64^\circ,\ \sqrt{4.25^2 + (19.9/\theta)^2}^\circ)$, $\theta$ in degrees} \\
& Geometry & $(C_f,C_{\mathrm{LC}})$ & \hr{$(3\ \mathrm{px},8.7^\circ)$} \\
Window
& Clones & $(N_w,N_k)$ & $(12,80)$ \\
\multirow{2}{*}{Tracks}
& Length/budget & $c$ & $c_{\text{min}} = 4$; at most 30 tracks per update \\
& Triangulation & -- & $8$ px reprojection threshold \\
\multirow{3}{*}{Loop}
& Warm-up & -- & 50 images \\
& Candidate & -- & Age $\geq2N_w$; $\Delta\theta\leq$ \hr{$45^\circ$} \\
& Acceptance & $N_m$ & $N_m\geq100$; NIS $\leq40$ \\
Covariance inflation & -- & $\lambda_\rho$ & 1.006 \\
\bottomrule
\end{tabular}

\label{tab:simulation_parameters}
\end{table*}

The original SPE3R dataset \cite{park_2024_scitech_spe3r} includes 1,000 synthetically generated images per real satellite model along with the associated masks and poses. 
The dataset contains 64 spacecraft models, with 7 of them being reserved for testing only and the rest for training. 
The original SPE3R dataset is rendered using the Unreal Engine 5 at a resolution of $256 \times 256$ px. 

SPE3R+ re-renders SPE3R in grayscale at $1920 \times 1200$ px using an OpenGL renderer \cite{bell2025event}.
A sample set of images are shown in \autoref{fig:spe3r_comparison} for comparison.
In addition to the original 64 models, 39 new models are introduced for a grand total of 103 spacecraft of which 11 are used for testing only. 
These spacecraft are rendered at varying distances such that their angular size in the image plane occupies a similar distribution for all of them regardless of their true size. 
5,000 images per satellite are rendered, along with masks and the metric depth which is used to estimate the true feature matches needed to fine-tune LightGlue. 
Although no image noise is inserted, a single light source is modeled as sunlight which varies in its incidence angle to provide a varied set of illumination conditions. 

The dataset only contains single images, and so to create image pairs for each satellite, images are compared against every other image in SPE3R+.
A pair is kept only if the viewing direction, measured as the angle between the camera boresights in the target frame, changes by less than $45^\circ$. 
Of the pairs used for SON, $66\%$ have a total rotation below $30^\circ$ and $25\%$ between $30^\circ$ and $45^\circ$, and are split into $8$ bins of total rotation, each capped to balance the distribution.
The remaining $10\%$ have a larger rotation about the boresight, so that SON also sees large in-plane rotations. 
As a result, $10\%$ of the test pairs exceed $45^\circ$ of total rotation. 
The pairs used to fine-tune LightGlue additionally include an equal number of wide-viewpoint pairs (viewing-direction change of at least $30^\circ$, total rotation below $120^\circ$), which are needed for loop closure. 
For the satellites in the training set, $20\%$ of the images per satellite are held out for validation, and then pairs are determined within the training and validation. 
% \autoref{fig:dataset_distribution} shows the distribution of image pairs between the different splits for each satellite. 

\subsubsection{Rendered trajectories} 
In addition to the SPE3R+ dataset, the rendering is used during the MSCKF testing. 
Here a trajectory is input and the reference truth dynamics generate a discrete set of poses that can be fed to the renderer. 
These are then rendered, and the Kalman filter is evaluated against these renderings which have similar appearance as those in SPE3R+. 

\begin{table*}[t]
\centering
\caption{Monte Carlo campaign and evaluation configuration.}
\footnotesize{\setlength\tabcolsep{4.0pt}%
\renewcommand{\arraystretch}{1.14}
\setlength{\heavyrulewidth}{0.4pt}
\setlength{\lightrulewidth}{0.4pt}
\setlength{\cmidrulewidth}{0.4pt}
\begin{tabular}{r r l l}
\toprule
Group & Quantity & Symbol/criterion & Value \\
\midrule
\multirow{2}{*}{Campaign}
& Targets/trials & $(N_T,N_{\mathrm{MC}})$ & $(11,50)$ \\
& Horizon/window & $(T_{\mathrm{MC}},\mathcal I_{\mathrm{eval}})$
& $2P_{\mathrm{orb}}$; final $25\%$ \\
\midrule
\multirow{5}{*}{Truth}
& Attitude & $\mathbf q_{T/C,0}$ & $\mathcal U(\mathrm{SO}(3))$ \\
& Angular velocity & ${}^{T}\boldsymbol\omega_{T/C,0}$
& \makecell[l]{Direction $\sim\mathcal U(\mathbb S^2)$;\\magnitude $\sim\mathcal U(0.6,1.2)^\circ/\mathrm{s}$} \\
& Along-track ROE & $a\delta\lambda_0$
& $-\max(\mathcal U(0.5,1.5),1)\cdot\rho_{\text{nom}}$\\
& Remaining ROE & $a[\delta a,\delta e_x,\delta e_y,\delta i_x,\delta i_y]_0$
& \makecell[l]{\hr{$0$}, $\mathcal U(-1.5,1.5)\%$, $(1+\mathcal U(-1.5,1.5))\%$, \\ $\mathcal U(-1.5,1.5)\%$, $(-1+\mathcal U(-1.5,1.5))\%$ of $\rho_{\text{nom}}$} \\
& Nominal range & $\rho_{\text{nom}}$ & Per target, \autoref{tab:mc_per_satellite} \\
\midrule
\multirow{4}{*}{Prior}
& Reference state & $(\hat{\mathbf q}_{T/C,0},{}^{T}\hat{\boldsymbol\omega}_{T/C,0})$
& $(\mathbf I,\mathbf0)$ \\
& ROE error & $a\hat{\delta\boldsymbol\alpha}_0-a\delta\boldsymbol\alpha_0$
& $\mathcal N\left(0,\operatorname{diag}([15,30,15,15,15,15]\% \text{ of } \rho_{\text{nom}})^2\right)$ \\
& ROE norm (consider) & $\hat{s}$ & \hr{$s_0$} \\
& $1\sigma$ & $(\sigma_{\theta,0},\sigma_{\omega,0},\boldsymbol\sigma_{\delta \alpha,0}, \sigma_s)$
& \makecell[l]{\hr{$0.01^\circ$}, $5^\circ/\mathrm{s}$;\\$[30,60,30,30,30,30]\% \text{ of }\rho_{\text{nom}}$; $\left( \hat{d}_0^\top \Sigma_{\delta \alpha,0} \hat{d}_0 \right)^{1/2}$ } \\
\midrule
\multirow{3}{*}{Evaluation}
& Converged & $(\bar e_{\mathrm{att}},\lVert e_{\mathrm{ROE}}\rVert_{\mathrm{RMS}})$ & \hr{$(<25^\circ,<\sigma_{a \delta \lambda,0})$} \\
& Meets specification & $(\bar e_{\mathrm{att}},\lVert e_{\mathrm{ROE}}\rVert_{\mathrm{RMS}})$
& $(<15^\circ,<0.1\bar\rho)$ \\
& Statistics & -- & $p_{50}$, $p_{95}$; median NEES \hr{per DOF} \\
\bottomrule
\end{tabular}
}
\label{tab:monte_carlo_parameters}
\end{table*}

\subsubsection{SHIRT}
The SHIRT \cite{park2023spnukf} dataset contains two rendezvous trajectories (ROE1 and ROE2) about the Tango spacecraft from the PRISMA mission \cite{d_pose_2014}, each with 2,371 images at a 5 s cadence. 
Every trajectory is imaged in two domains: synthetic and lightbox.
The synthetic domain is much like the rendered trajectories described above, but includes Gaussian image noise. 
The lightbox domain contains images captured by a real camera within the TRON facility \cite{park2021tron} which reproduces space-like illumination and sensor effects. 
Tango is not a part of SPE3R+, and so it is a previously unseen target much like the held-out test set. 
SHIRT does not provide image masks, which are needed by this pipeline to create the bounding box, and so these are re-generated as in SPE3R+. 
\hr{The odometry front-end is additionally evaluated on noise-free re-renders of the SHIRT poses to measure the sim-to-real gap.} 

\subsection{Reference-truth propagation}

Reference truth servicer and target orbit and attitude trajectories are propagated jointly using \textit{S3} \cite{bell2025event, giralo2018development}. 
The specific orbit and dynamics models used are reported in \autoref{tab:simulation_parameters}.
Target inertias are computed from voxelized shapes assuming uniform density, and scaled using the reported launch masses. 

In addition to the moment of inertia, a coarse 3D model is required in order to calculate the torques associated with drag and Solar Radiation Pressure (SRP). 
A two-box decomposition (a central bus plus two symmetric thin solar panels) is fit to the calculated moment of inertia and bounding box, falling back to a single uniform cuboid if the fit fails.
The bulk drag and SRP areas are instead measured as the mean silhouette area of the target mesh over random orientations because a box decomposition cannot represent panels shading the bus.

\subsection{Fine-tuning LightGlue}
The original authors of LightGlue \cite{lindenberger_lightglue_2023} first train the model using SuperPoint features \cite{detone_superpoint_2018} on homographies sampled from 1 million images \cite{radenovic2018revisiting}.
The model is then fine-tuned with the full MegaDepth dataset \cite{li2018megadepth}. 

LightGlue is further fine-tuned on SPE3R+. 
The particular training parameters are provided in \autoref{tab:training_parameters}. 
LightGlue and SON use the same train, validation, and test image splits of SPE3R+, but different pair sets as per \autoref{tab:training_parameters}.

\subsection{Training SON}
To accelerate the training, SuperPoint and LightGlue are run as a pre-processing step on all image pairs in SPE3R+.
\autoref{tab:training_parameters} lists the baseline SON architecture and training settings. 

% \subsection{Kalman filter}
% The MSCKF used in this work is initialized using a noisy reference truth for the ROE portion of the state, under the assumption that in operations some previous method has been used when the target was further away. 
% The arbitrary body orientation on the other hand is defined such that the initial attitude of the target is aligned with the camera frame at that time. 
% The relative angular velocity is initialized at $0^\circ/s$. 
% Table~\ref{tab:simulation_parameters} consolidates the simulation, control, and MSCKF parameters used throughout the end-to-end evaluation.
% \autoref{tab:monte_carlo_parameters} on the other hand, lists the Monte Carlo sampling distributions, filter prior, and evaluation criteria.

\begin{table*}[t]
\centering
\caption{The performance of the LightGlue feature matcher (using SuperPoint features) is compared with different weights: the original weights released by Lindenberger et al., and the ones fine-tuned using SPE3R+. On HPatches Precision (P) and homography estimation accuracy using RANSAC and DLT is compared. On MegaDepth1500 and SPE3R+ match Precision (P) and Recall (R) as well as the pose accuracy from RANSAC is compared. On SPE3R+ the number of matches (\#M) and the number of RANSAC inliers (Inl) are also reported.}
\footnotesize
\setlength\tabcolsep{3.0pt}%
\renewcommand{\arraystretch}{1.3}
\setlength{\heavyrulewidth}{0.4pt}
\setlength{\lightrulewidth}{0.4pt}
\setlength{\cmidrulewidth}{0.4pt}
\begin{tabular}{c ccccc ccc ccccc}
\toprule
& \multicolumn{5}{c}{HPatches} & \multicolumn{3}{c}{MegaDepth1500} & \multicolumn{5}{c}{SPE3R+ (test)} \\
& P & \multicolumn{2}{c}{AUC - RANSAC} & \multicolumn{2}{c}{AUC - DLT} & P & R & RANSAC AUC & P & R & \#M & Inl & RANSAC AUC \\
Weights & & @1px & @5px & @1px & @5px & & & 5° / 10° / 20° & & & & & 5° / 10° / 20° \\
\midrule
Original & \multirow{2}{*}{\textbf{94.1}} & \multirow{2}{*}{\textbf{34.5}} & \multirow{2}{*}{\textbf{68.5}} & \multirow{2}{*}{\textbf{35.2}} & \multirow{2}{*}{\textbf{77.6}} & \multirow{2}{*}{\textbf{83.5}} & \multirow{2}{*}{\textbf{90.4}} & \multirow{2}{*}{\textbf{50.3} / \textbf{67.6} / \textbf{80.5}} & 28.7 & 32.7 & 89 & 27 & 0.0 / 0.2 / 0.8 \\
Original + BBox &  &  &  &  &  &  &  &  & 32.9 & 15.7 & 346 & 87 & 0.5 / 1.6 / 4.0 \\
\midrule
Fine-Tuned + BBox & 88.2 & 32.3 & 64.7 & 30.3 & 70.5 & 62.1 & 58.5 & 31.6 / 46.2 / 57.7 & \textbf{59.5} & \textbf{49.1} & \textbf{694} & \textbf{274} & \textbf{3.8} / \textbf{8.6} / \textbf{17.0} \\
\bottomrule
\end{tabular}
\label{tab:fine_tuning_results}
\end{table*}

\subsection{Monte Carlo analysis}
The full navigation architecture described in \autoref{fig:overall_pipeline} is evaluated on the 11 spacecraft reserved for the SPE3R+ test split. 
For each of these targets, 50 Monte Carlo (MC) trials are propagated for two orbit periods using the simulation parameters described in \autoref{tab:simulation_parameters}. 
The relative orbit for each trial is scaled to the standoff chosen for that target.
\hr{The number of trials is limited by the rendering cost of each trajectory, and is sufficient to identify failure modes and compare design choices, although tighter bounds on rare failures would require more trials.}
Standoffs are set per target by hand, where being closer improves \hr{image resolution}, but farther improves safety. 
The ROE of each trial are drawn from the distribution in \autoref{tab:monte_carlo_parameters} expressed as fractions of the target's nominal range $\rho_{\text{nom}}$, shown in \autoref{tab:mc_per_satellite}.

The filter fixes the arbitrary target-frame attitude gauge by setting the initial relative attitude to identity \hr{with very small uncertainty, since the target frame is defined as the camera frame at $t_0$ \cite{zhang2018comparison},} and initializes the relative angular velocity to $\mathbf{0}$. \hr{A non-zero prior on this unobservable rotation would otherwise be slowly and spuriously reduced by the EKF linearization, making the filter over-confident.}
The initial ROE estimate is obtained by perturbing the reference truth value according to the distribution provided in \autoref{tab:monte_carlo_parameters}.

Metrics for each MC trial are evaluated over the final $25\%$ of each trajectory. 
Attitude errors are computed after the constant-rotation alignment described in \autoref{sec:msckf} \cite{zhang2018tutorial}.
A trial is classified as converged according to the criterion reported in \autoref{tab:monte_carlo_parameters}.
A converged trial meets specification when its mean attitude error is below $15^\circ$ and its total ROE root-mean-square error\hr{, after scaling each step's estimate by the least-squares factor that best matches the truth,} is below $0.1\bar{\rho}$, where $\bar{\rho}$ is the mean target--servicer separation.
Across trials, estimation errors are reported using the 50th and 95th percentiles, denoted by $p_{50}$ and $p_{95}$.
Normalized Estimation Error Squared (NEES) is reported as the median of the per-trial evaluation window means\hr{, normalized per degree of freedom so that a consistent filter gives 1. The attitude NEES is computed with the unobservable attitude gauge, a common rotation of all attitude states, projected out of both the error and the covariance.
} 

\subsection{Inference analysis}
Inference time and memory are measured per pipeline stage, along with network sizes, on an NVIDIA Jetson AGX Orin 64 GB.

%%%%%%%%%%%%%%%%%%%%%%%%%%%%%%%%%%%%%%
\section{Results} \label{sec:results}
%%%%%%%%%%%%%%%%%%%%%%%%%%%%%%%%%%%%%%
The results and their explanation are covered for each section of this work individually below. 
The overall pipeline and its performance is discussed in a final section.

\subsection{Fine-tuning LightGlue}
The performance of LightGlue is evaluated in three different cases on three different datasets in \autoref{tab:fine_tuning_results}. 
The first row of this table represents the original weights of LightGlue released by the authors and performs well on its target regime as shown by its state-of-the-art performance on HPatches and MegaDepth1500. 
When turning to the SPE3R+ test set, the performance falls precipitously: pose AUC at $20^\circ$ decreases from $80.5$ to $0.8$. 
This makes this particular odometry measurement insufficient for the downstream estimator when compared to the terrestrial baseline on MegaDepth1500. 

Some of these images can contain limited texture, and commonly bounding boxes are employed to crop the usually black space background. 
This is only done for SPE3R+, and so the second row shares the same metrics on HPatches and MegaDepth1500 as the first row. 
With the addition of the bounding box, the performance more than quadruples in the RANSAC AUC metrics likely due to the increased number of matches identified, but the performance remains insufficient for odometry estimation. 
Recall decreases likely because as the crop is applied, more reference truth matches are available, and so the denominator grows leading to the decrease.

\begin{figure}[htbp]
    \centering
    \includegraphics[width=\linewidth]{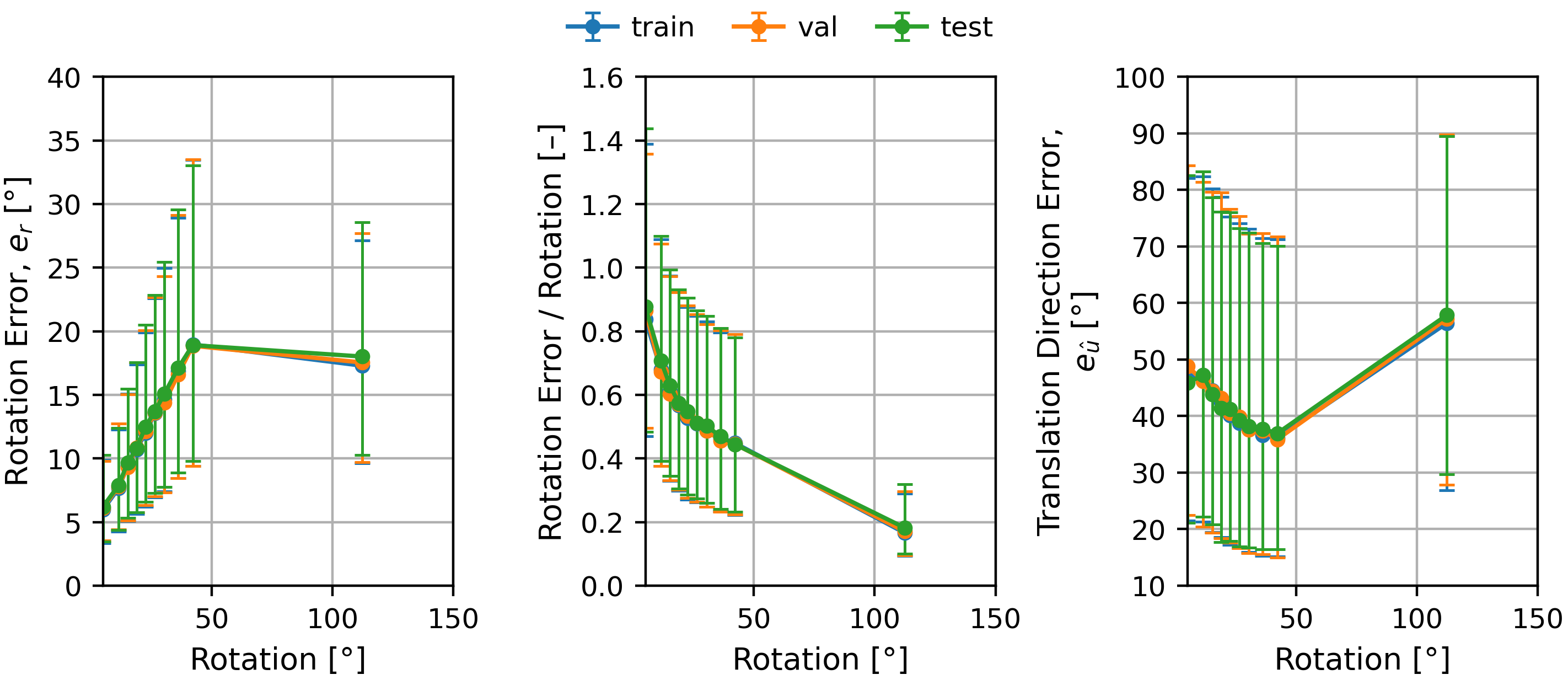}
    \caption{Performance of fine-tuned LightGlue and the classical odometry estimator on SPE3R+ data splits.}
    \label{fig:essmat_errors}
\end{figure}

The last row shows the performance of LightGlue when it is fine-tuned on SPE3R+ and bounding boxes are provided. 
RANSAC AUC increases by factors of approximately $4$--$8$ across the three thresholds, accompanied by an increase in the number of matches and RANSAC inliers.
There is some loss of generality in these new weights, as the performance on terrestrial image pairs as seen with HPatches and MegaDepth1500 does decrease. 
Since the decrease is not as significant as the increase in performance in SPE3R+, and the focus of this work is purely spacecraft pose estimation this is not a significant concern. 

Even with these increases, the performance on spacecraft images as seen by the RANSAC AUC metrics is still significantly worse than on terrestrial images, and so this motivates the use of SON. 

\begin{table*}
    \centering
    \caption{Median rotation and translation unit vector errors for the different data splits of SPE3R+ from the classical baseline of \Cref{sec:odometry_estimation_from_feature_matches} (essential matrix with USAC / MAGSAC++) and different SON architectures. Interquartile range shown through the super and subscript.}
    \footnotesize
    \setlength\tabcolsep{4.0pt}%
\setlength{\heavyrulewidth}{0.4pt}
\setlength{\lightrulewidth}{0.4pt}
\setlength{\cmidrulewidth}{0.1pt}
\renewcommand{\arraystretch}{1.7}
\begin{tabular}{rr  c c  c c  c c}
    \toprule
    \multicolumn{2}{c}{} & \multicolumn{2}{c}{Train} & \multicolumn{2}{c}{Val} & \multicolumn{2}{c}{Test}  \\
    & & $\epsilon_{rot}$ $[^\circ]$ & $\epsilon_{\hat{u}}$ $[^\circ]$ & $\epsilon_{rot}$ $[^\circ]$ & $\epsilon_{\hat{u}}$ $[^\circ]$ & $\epsilon_{rot}$ $[^\circ]$ & $\epsilon_{\hat{u}}$ $[^\circ]$ \\
    \midrule
    \multicolumn{2}{r}{Baseline} & $11.4^{+9.4}_{-5.5}$ & $42.8^{+35.8}_{-24.3}$ & $13.1^{+9.6}_{-6.3}$ & $41.8^{+35.8}_{-24.0}$ & $11.0^{+9.1}_{-5.3}$ & $43.5^{+35.2}_{-24.0}$ \\
    \midrule
    \multirow{4}{*}{\parbox{2.2em}{\centering\rotatebox{90}{SON}}} 
      & MLP Only & $13.1^{+7.3}_{-5.0}$ & $37.7^{+25.5}_{-16.2}$ & $15.4^{+6.6}_{-5.2}$ & $37.0^{+26.5}_{-16.2}$ & $12.6^{+7.8}_{-4.9}$ & $41.4^{+26.5}_{-17.7}$ \\
      & 1 Layer, Rand Init. & $10.1^{+5.2}_{-3.6}$ & $21.5^{+12.5}_{-8.5}$ & $12.1^{+5.4}_{-4.0}$ & $23.3^{+15.7}_{-9.5}$ & $11.0^{+6.2}_{-4.0}$ & $29.0^{+20.4}_{-12.2}$ \\
      & 1 Layer & $7.4^{+4.1}_{-2.8}$ & $17.1^{+9.5}_{-6.7}$ & $9.0^{+4.6}_{-3.2}$ & $18.7^{+12.0}_{-7.5}$ & $8.5^{+5.3}_{-3.3}$ & $22.9^{+16.0}_{-9.4}$ \\
      & 9 Layers & $\mathbf{2.3^{+1.9}_{-1.0}}$ & $\mathbf{2.8^{+1.8}_{-1.2}}$ & $\mathbf{2.9^{+2.3}_{-1.3}}$ & $\mathbf{3.1^{+2.5}_{-1.4}}$ & $\mathbf{2.9^{+2.6}_{-1.3}}$ & $\mathbf{4.1^{+3.6}_{-1.9}}$ \\
    \bottomrule
\end{tabular}
    \label{tab:son_performance}
\end{table*}

\subsection{Classical odometry estimation}
For additional context, \autoref{fig:essmat_errors} reports the rotation and translation-direction errors underlying the SPE3R+ RANSAC AUC values as functions of the reference rotation.
The results are binned into equal-count bins (so regions with more samples have more bins), and the median and interquartile range for these bins are shown.
The results are split into three sub-plots, the first shows the rotation error with respect to the rotation.
As expected, the rotation error increases as the rotation increases, however as is seen by the second plot this increase in error is slower than the rotation increase. 
This is expected: larger rotations will lead to larger rotation errors, but small rotations are difficult to pin down at all, and so yield a low signal to noise ratio. 
The final sub-plot at the far right shows the translation error as the angle between the true and estimated translation unit direction vectors. 
Here the errors remain around $35$--$60^\circ$ regardless of the rotation. 

The errors are further broken down by the SPE3R+ data splits. 
Similar split-wise performance indicates limited sensitivity to whether a spacecraft geometry appears during LightGlue fine-tuning.
Nevertheless, as discussed in the previous section this performance is still too poor for the precise pose estimation desired in this work. 

\subsection{SON results}

\begin{figure}[htbp]
    \centering
    \includegraphics[width=\linewidth]{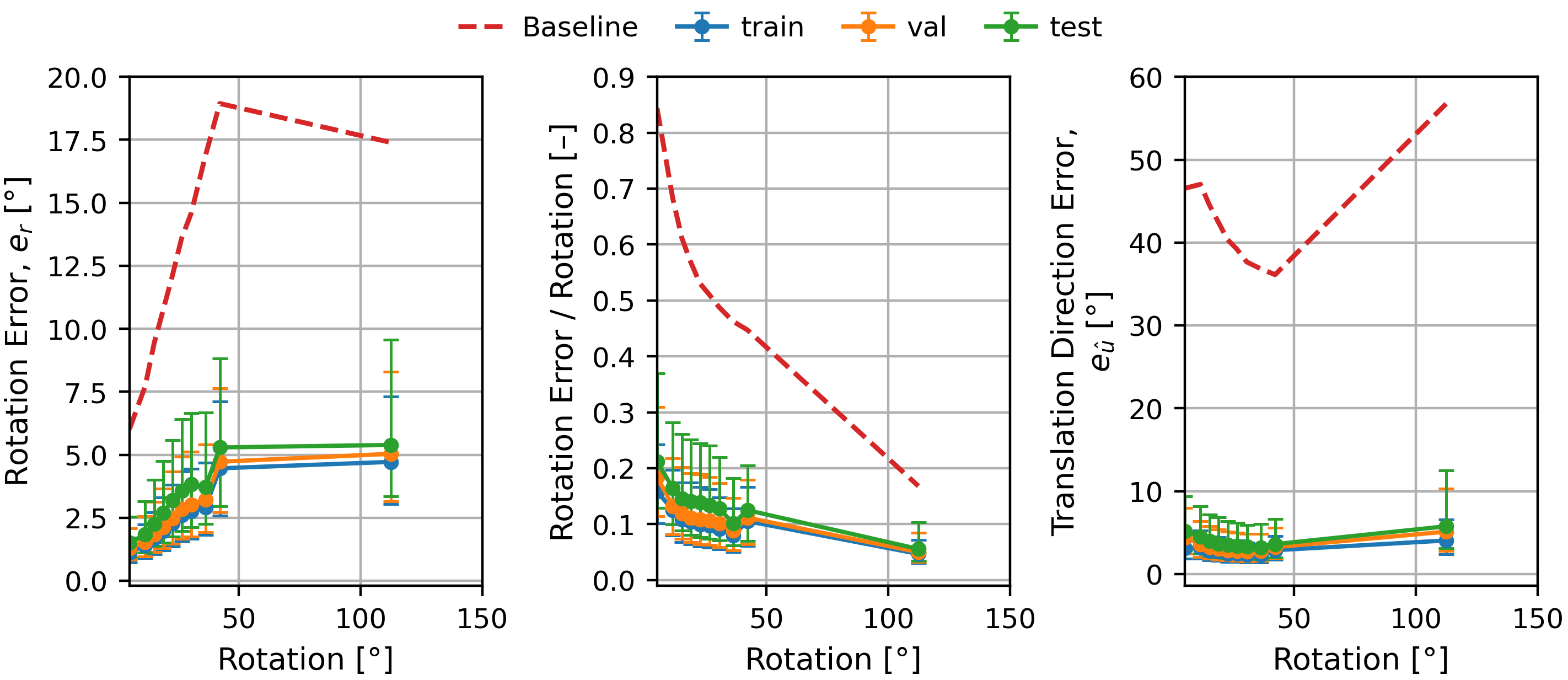}
    \caption{Performance of fine-tuned LightGlue and SON on SPE3R+ data splits.}
    \label{fig:son_errors}
\end{figure}

The performance of SON on SPE3R+ is now shown in \autoref{fig:son_errors}, where now the median performance of the classical odometry estimator is shown as a dashed red line. 
When compared against the baseline, SON reduces errors relative to it and as will be demonstrated in further results this increase in performance \hr{substantially improves the attitude estimate within the filter}. 
Although test errors exceed training errors, SON still reduces both errors substantially relative to the classical baseline on held-out spacecraft.
Even with this performance increase, low rotation image pairs still pose a signal-to-noise ratio problem as the amount of error is large with respect to the actual rotation. 

SON's advantage in rotation estimation over the classical baseline is larger on SPE3R+ image pairs than on consecutive trajectory images, as will be seen in further results.
This is because SPE3R+ draws images with random sun illumination, whereas odometry in the filter is run between image pairs that are close in time, so the illumination is consistent. 
SON's advantage will translate to better loop closure performance though, where the images may be from distinctly different times. 

To validate the choice of architecture, a small ablation is performed, with the results shown in \autoref{tab:son_performance}. 
The first row of this table again repeats the baseline results for reference, and the rest represent the ablation on SON. 
The last row of this table then shows the performance of SON at its full specification with 9 transformer layers. 
This is the same SON that is used to generate the results in \autoref{fig:son_errors}.
In these results the rotation error on the test set is lower than or equal to that on the validation set.
The reason for this is that the validation pairs are formed from only $20\%$ of the held out images, meaning there are fewer images to choose from. 
This means there are fewer low rotation pairs, and so the rotation error is expected to be larger on average (since the average rotation is larger). 
The next row up is the first ablation, where the question of how many transformer layers are needed is asked. 
With one layer, both networks generalize similarly (the difference between the train and test set performance is similar), but nine layers reduce the median errors by factors of about 3 and 6 in the rotation and translation direction respectively. 

The next ablation asks whether the weights from LightGlue are beneficial. 
The original philosophy of this network is that good feature matchers can be good odometry estimators.
This ablation therefore tests whether the network architecture is the reason for this additional performance, or if the weights themselves also have some effect. 
Comparing the third and fourth rows the only difference is that in the third row the weights of the SON transformers are initialized randomly instead of with the weights of LightGlue (as is done for all other cases). 
The LightGlue initialization reduces the median errors by about $20$--$25$\%, but this ablation does not identify its cause.

The final ablation tests this original philosophy and removes the transformers entirely, instead opting to only use the MLP. 
Here the performance is either on par or worse than the baseline, showing that at least for the selected architectures the transformers are essential to the performance of the network.

\autoref{tab:son_performance} shows that the transformers provide a large boost to training performance, and going from 1 to 9 layers increases this performance without significant effects on the network's capacity to generalize. 

\subsection{Performance on SHIRT}
\begin{table*}
    \centering
    \caption{Performance of SON and the classical baseline (\Cref{sec:odometry_estimation_from_feature_matches}) on sequential images from the SHIRT \cite{park2023spnukf} dataset. Noise-free image domain is a re-rendered version of SHIRT using the same renderer as SPE3R+. Errors (median with IQR) are reported for image pairs on which both estimators provide a measurement, which requires at least 8 feature matches. The proportion of failures of the baseline method on the pairs with at least 8 feature matches is also reported (SON always provides a measurement).}
    \footnotesize
    \setlength\tabcolsep{4.0pt}%
\setlength{\heavyrulewidth}{0.4pt}
\setlength{\lightrulewidth}{0.4pt}
\setlength{\cmidrulewidth}{0.4pt}
\renewcommand{\arraystretch}{1.7}
\begin{tabular}{ll cc cc cc}
\toprule
& & \multicolumn{1}{c}{Pairs [\%]} & \multicolumn{1}{c}{Baseline} & \multicolumn{2}{c}{SON} & \multicolumn{2}{c}{Baseline} \\
\cmidrule(lr){5-6} \cmidrule(lr){7-8}
Trajectory & Domain & $N_m \geq 8$ & fails [\%] & $\epsilon_{rot}$ $[^\circ]$ & $\epsilon_{\hat{u}}$ $[^\circ]$ & $\epsilon_{rot}$ $[^\circ]$ & $\epsilon_{\hat{u}}$ $[^\circ]$ \\
\midrule
\multirow{3}{*}{ROE1} & Noise-free & 100 & 0 & $\mathbf{1.1^{+0.7}_{-0.4}}$ & $\mathbf{2.8^{+1.6}_{-1.1}}$ & $1.2^{+0.7}_{-0.5}$ & $10.9^{+10.4}_{-5.9}$ \\
& Synthetic & 69 & 38 & $7.5^{+13.1}_{-4.4}$ & $\mathbf{10.6^{+23.6}_{-6.1}}$ & $\mathbf{3.3^{+1.6}_{-1.8}}$ & $45.6^{+40.4}_{-25.7}$ \\
& Lightbox & 50 & 17 & $4.8^{+8.3}_{-3.3}$ & $\mathbf{12.5^{+143.2}_{-7.5}}$ & $\mathbf{4.0^{+1.0}_{-1.8}}$ & $58.9^{+31.4}_{-27.3}$ \\
\midrule
\multirow{3}{*}{ROE2} & Noise-free & 100 & 1 & $\mathbf{0.9^{+0.4}_{-0.3}}$ & $\mathbf{5.0^{+3.8}_{-2.2}}$ & $\mathbf{0.9^{+0.7}_{-0.4}}$ & $24.4^{+20.4}_{-12.5}$ \\
& Synthetic & 91 & 14 & $2.1^{+1.7}_{-0.9}$ & $\mathbf{6.5^{+6.5}_{-2.9}}$ & $\mathbf{1.3^{+1.1}_{-0.6}}$ & $36.1^{+30.7}_{-20.3}$ \\
& Lightbox & 97 & 9 & $2.2^{+2.1}_{-1.0}$ & $\mathbf{11.3^{+29.4}_{-5.7}}$ & $\mathbf{2.0^{+1.3}_{-0.9}}$ & $58.1^{+32.7}_{-28.3}$ \\
\bottomrule
\end{tabular}
    \label{tab:shirt_odometry}
\end{table*}

Odometry estimation is now compared between three image domains in \autoref{tab:shirt_odometry}.
In the noise-free domain for both trajectories the advantage of SON is minimal in rotation estimation, but it does provide a $4$--$5$ times lower translation unit direction error. 
Here, both methods provide a measurement for almost every pair. 
Once image noise is added in the SHIRT synthetic images, the number of image pairs with fewer than 8 feature matches increases to $31\%$ for ROE1 and $9\%$ for ROE2. 
This causes increases in error in both rotation and translation unit direction, and SON’s rotation error now exceeds the baseline’s, while its translation unit direction error remains $4$--$6$ times lower.
Hidden from these error metrics is the fact that the baseline fails to report a measurement significantly more often (for more than a third of the image pairs in the case of ROE1), which would lead to significant measurement dropout in a filter.
Finally, when real images from the lightbox domain are introduced the results are a mixed bag for SON and the baseline. 
Beginning with the latter, its errors increase, but the number of failures decreases, though on ROE1 only half the pairs reach 8 matches. 
For SON, the rotation error is slightly above the baseline’s, $4.8^\circ$ against $4.0^\circ$ on ROE1 and $2.2^\circ$ against $2.0^\circ$ on ROE2, and translation unit direction error increases by a factor of $1.2$--$1.7$ relative to the synthetic domain. 
Additionally, on ROE1 over a quarter of the image pairs have a reversed translation unit direction as seen by the upper quartile at $156^\circ$. 

\hr{Since these lightbox errors exceed the SON noise model used in the filter, \autoref{tab:simulation_parameters}, closing this sim-to-real gap is a prerequisite for flight. 
Except for the translation direction on ROE2, most of the degradation already appears between the noise-free and synthetic domains, which differ mainly by image noise that is absent from SPE3R+. 
Adding sensor noise to the training images of LightGlue and SON, together with retuning the noise model, is therefore expected to recover much of the gap and is the focus of future work.}

\subsection{Monte Carlo results}

\begin{table*}
    \centering
    \caption{Performance of the whole pipeline on each of the SPE3R+ test-spacecraft on the Monte Carlo trajectories. $\rho_{nom}$ is the designed standoff distance, and $\bar{\rho}$ is the mean separation over the entire trajectory over the 50 trials. }
    \footnotesize
    \setlength\tabcolsep{3.0pt}%
\renewcommand{\arraystretch}{1.3}
\setlength{\heavyrulewidth}{0.4pt}
\setlength{\lightrulewidth}{0.4pt}
\setlength{\cmidrulewidth}{0.4pt}
\begin{tabular}{r cccc ccc ccc ccc}
\toprule
& & & & & \multicolumn{3}{c}{Attitude $[^\circ]$} & \multicolumn{3}{c}{ROE $[\%]$} & \multicolumn{3}{c}{$\omega$ $[^\circ/s]$} \\
Satellite name & Converged $[\%]$ & Meets Spec. $[\%]$ & $\rho_{nom}$ $[m]$ & $\bar{\rho}$ $[m]$ & p50 & p95 & NEES & p50 & p95 & NEES & p50 & p95 & NEES \\
\midrule
AIM & 100 & 100 & 10.0 & 10.4 & 2.7 & 6.8 & 2.3 & 0.9 & 1.6 & 2.2 & 0.12 & 0.20 & 0.08 \\
Aquarius & 100 & 100 & 10.0 & 12.4 & 2.4 & 4.4 & 3.1 & 1.2 & 1.9 & 2.6 & 0.09 & 0.13 & 0.04 \\
CYGNSS & 100 & 100 & 5.0 & 5.8 & 3.7 & 12.1 & 3.3 & 1.3 & 3.9 & 1.5 & 0.15 & 0.25 & 0.13 \\
EO-1 & 100 & 100 & 10.0 & 12.4 & 3.1 & 6.0 & 3.4 & 1.1 & 1.8 & 2.0 & 0.11 & 0.15 & 0.07 \\
Europa Orbiter & 100 & 98 & 20.0 & 22.8 & 2.4 & 3.2 & 2.0 & 0.6 & 0.9 & 1.7 & 0.10 & 0.14 & 0.06 \\
Galileo & 100 & 100 & 30.0 & 34.6 & 2.7 & 5.7 & 3.0 & 0.6 & 1.0 & 2.6 & 0.10 & 0.13 & 0.06 \\
Herschel & 100 & 100 & 20.0 & 23.5 & 2.0 & 2.8 & 2.4 & 0.7 & 1.2 & 2.0 & 0.08 & 0.11 & 0.04 \\
Sentinel-6 & 100 & 98 & 15.0 & 17.8 & 2.5 & 6.4 & 3.2 & 0.9 & 1.4 & 1.7 & 0.10 & 0.15 & 0.05 \\
SMART-1 & 32 & 18 & 25.0 & 27.4 & 45.4 & 118.2 & 132.5 & 0.7 & 2.4 & 5.2 & 0.66 & 1.25 & 1.36 \\
Spitzer & 100 & 100 & 10.0 & 12.0 & 2.4 & 6.5 & 2.7 & 1.1 & 1.9 & 1.9 & 0.10 & 0.15 & 0.05 \\
SSL-1300 & 100 & 100 & 25.0 & 28.1 & 2.3 & 3.9 & 1.9 & 0.6 & 1.0 & 2.6 & 0.13 & 0.17 & 0.09 \\
\midrule
All (excl. SMART-1) & 100.0 & 99.6 & - & 18.0 & 2.5 & 6.3 & 2.8 & 0.9 & 1.8 & 2.1 & 0.10 & 0.17 & 0.06 \\
\bottomrule
\end{tabular}
    \label{tab:mc_per_satellite}
\end{table*}

\begin{figure}[htbp]
    \centering
    \includegraphics[width=0.6\linewidth]{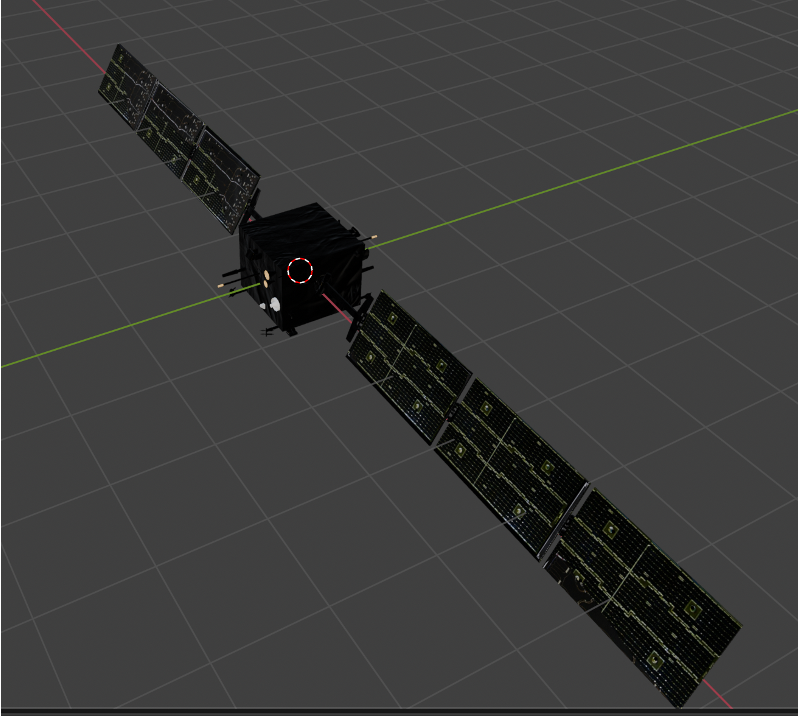}\\[1ex]
    \includegraphics[width=0.8\linewidth]{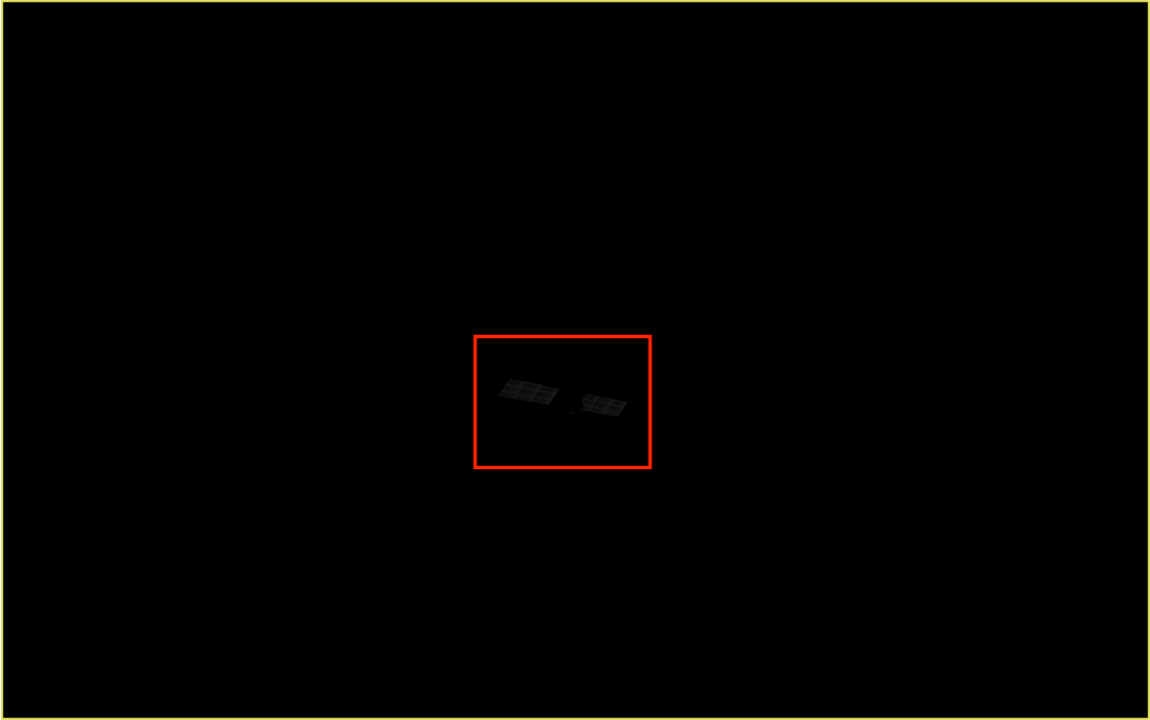}
    \caption{View within Blender of SMART-1 (top) and a sample image from SPE3R+ of this spacecraft (bottom). A red bounding box is used to highlight the target in the image. \vspace{-0.5cm}}
    \label{fig:smart1_example}
\end{figure}

The performance of the whole pipeline is evaluated through a set of Monte Carlo experiments on the 11 satellites within the SPE3R+ test set. 
These satellites are unseen during training, and so they represent the exact conditions this pipeline is meant to tackle. 
The results of these Monte Carlo experiments are shown in \autoref{tab:mc_per_satellite}, and are broken down per satellite. 
The first two columns are simple binary conditions for each experiment, and at a glance provide an expectation of performance. 
In these first columns it is already clear that there is an outlier in SMART-1, which rarely converges or meets the specification. 
The reason for this is clear when evaluating images from this satellite: it is covered in a dark mylar which hardly reflects any light, making the images severely underexposed. 
This satellite is therefore considered an edge case for this pipeline. 
The satellite model of SMART-1 used to develop SPE3R+ and a sample image are shown in \autoref{fig:smart1_example}.

\begin{table*}
    \centering
    \caption{Leave-one-out ablation. Each row removes the stated portion from the finished filter in \autoref{fig:overall_pipeline}. Statistics are given for the 200 trials per row (20 trials for the 10 targets, with SMART-1 removed).}
    \footnotesize
    \setlength\tabcolsep{3.0pt}%
\renewcommand{\arraystretch}{1.3}
\setlength{\heavyrulewidth}{0.4pt}
\setlength{\lightrulewidth}{0.4pt}
\setlength{\cmidrulewidth}{0.4pt}
\begin{tabular}{r cc ccc ccc ccc}
\toprule
& & & \multicolumn{3}{c}{Attitude $[^\circ]$} & \multicolumn{3}{c}{ROE $[\%]$} & \multicolumn{3}{c}{$\omega$ $[^\circ/s]$} \\
Variant & Converged $[\%]$ & Meets Spec. $[\%]$ & p50 & p95 & NEES & p50 & p95 & NEES & p50 & p95 & NEES \\
\midrule
Full pipeline & 100.0 & 99.5 & 2.5 & 6.1 & 2.8 & 0.9 & 1.8 & 2.1 & 0.11 & 0.18 & 0.07 \\
\midrule
SON $\rightarrow$ Baseline & 99.5 & 98.5 & 4.3 & 9.4 & 6.4 & 0.9 & 1.8 & 2.2 & 0.13 & 0.24 & 0.10 \\
No feature tracks & 92.5 & 67.0 & 10.0 & 36.1 & 5.1 & 7.0 & 14.9 & 1.7 & 0.26 & 0.53 & 0.30 \\
No loop closure & 88.5 & 70.0 & 10.9 & 31.9 & 0.7 & 0.9 & 1.9 & 2.2 & 0.19 & 0.46 & 0.06 \\
No covariance inflation & 98.5 & 97.0 & 2.6 & 7.1 & 2.7 & 0.9 & 2.4 & 6.6 & 0.11 & 0.21 & 0.07 \\
\bottomrule
\end{tabular}
    \label{tab:mc_breakdown_analysis}
\end{table*}

The following columns provide the attitude\hr{, scale-aligned ROE, and angular velocity} errors along with the NEES. 
\hr{With the range not estimated, the scale-aligned ROE errors are similar fractions of the range for all targets at median errors of $0.6\%$--$1.3\%$.}
The median attitude error, below \hr{$4^\circ$} for every target except SMART-1, is only weakly affected by the distance to or the size of the target. 
\hr{The tail of the distribution is not: the 95th percentile is largest for the closest targets ($12.1^\circ$ for CYGNSS at $5$ m, $4–7^\circ$ at $10–15$ m), and below $6^\circ$ from $20$ m onward.
Qualitatively, attitude errors grow when little of the target’s geometry changes in view, for example when the tumble axis lies close to the line of sight or the target is seen edge-on, and symmetric targets can produce ambiguous loop closures, which the rotation gate and NIS test are designed to reject.}
The NEES \hr{in \autoref{tab:mc_per_satellite} is normalized per degree of freedom, so a consistent filter gives $1$.}
\hr{The filters are over-confident in both attitude and ROE, with median values of $2.8$ and $2.1$ respectively, consistent with the correlated SON and feature-track errors discussed in \autoref{sec:msckf}.
The angular velocity is under-confident on the other hand, as its process noise absorbs the torque-free motion that is omitted by the kinematic model in \autoref{eqn:kinematic_ang_vel}.}

The final row of \autoref{tab:mc_per_satellite} summarizes the remaining $10$ targets. 
This is done because the errors vary widely for each satellite due to the range, and so large satellites (with large ranges) such as SSL-1300 and Galileo would dominate. 
Across these test satellites, the attitude stays below \hr{$6.3^\circ$} and the ROE errors stay below \hr{$1.8$\%} of the total range in more than $95\%$ of cases. 
Only two of the $500$ trials miss the specification, with mean attitude errors of $18^\circ$ and $17^\circ$ while still converging. 

\hr{As discussed in \Cref{sec:sota}, no published method addresses the same problem as this work with a shared benchmark, so direct comparisons are infeasible. 
The closest results are single-image estimators for unknown targets with errors in the tens of degrees \cite{park_2024_scitech_spe3r, mathihalli2024dreamsat, uwumukiza2026dreamsat}, and monocular SLAM that does not recover the scale and is validated over short or 3-DOF trajectories \cite{dor_orb-slam_2018, augenstein2011monocular, ticozzi2024factor}.
The classical front end of \Cref{sec:odometry_estimation_from_feature_matches} is therefore used as the benchmark, substituted into the same filter in \autoref{tab:mc_breakdown_analysis}.}

% \hr{Attitude failures concentrate in trials where the target spins near the line of sight. The same side then stays in view, loop closures are available at almost every step, and the per-frame parallax is small. Four of the five failures  share the same draw of initial conditions (so the same spin) on different targets.}

This pipeline as a whole includes a lot of moving pieces and hyperparameters, so a clear question is whether each of these pieces adds value to the end result. 
\autoref{tab:mc_breakdown_analysis} answers this by removing one piece at a time from the finished filter and running 20 trials for the 10 targets. 
The first row is the full pipeline, and the rows below it remove a single mechanism.

\begin{figure*}[t]
    \centering
    \includegraphics[width=\linewidth]{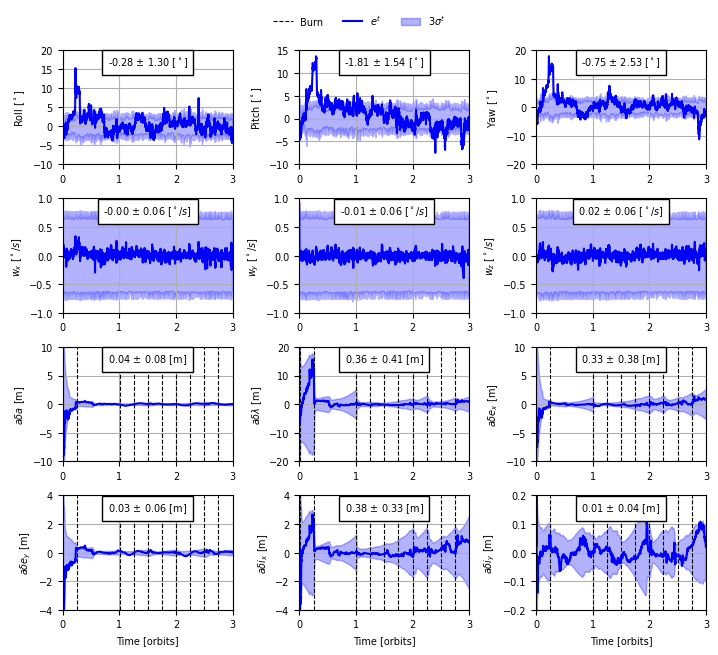}
    \caption{Error of the nominal state along with the $\pm 3 \sigma$ bound for a representative Monte Carlo trial with the Herschel spacecraft. Boxed quantity represents the mean and standard deviation of the error for the last quarter of the trajectory. Vertical lines represent orbit maneuvers.}
    \label{fig:mission_navigation}
\end{figure*}

\hr{Replacing SON with the classical estimator raises the median attitude error from $2.5^\circ$ to $4.3^\circ$, and the 95th percentile grows from $6.1^\circ$ to $9.4^\circ$, showing the attitude solution is less robust. 
This in turn lowers the rate of trials meeting specification, while ROE accuracy is unchanged.}
\hr{Its rotation, translation-direction, and loop-closure noise models are refit to its own errors on the same trajectories. 
Its translation-direction error is $3$--$4$ times larger.}
Loop closure uses the same classical solver \hr{and gate}, which is where SON's robustness to illumination change is important. 
\hr{Across these trials the classical solver fails to return a loop-closure solution about 12 times as often as SON, while the odometry rotation errors of the two are nearly identical. 
SON’s attitude advantage therefore comes from loop closure, and its translation-direction advantage does not reach the ROE likely because the feature tracks constrain it.}

The remaining rows each degrade a different portion of the state estimate. 
Without the feature tracks, only $67$\% of trials meet the specification, and the median errors grow four-fold in attitude and eight-fold in ROE.
Without loop closure the ROE is unchanged and its additional failures come from attitude errors only.
Here the median attitude error goes from \hr{$2.5^\circ$ to $10.9^\circ$}, showing that the feature tracks' short baseline constraint is insufficient to ensure the attitude remains self-consistent. 
Finally, without covariance inflation the median errors barely move, though the 95th-percentile ROE error grows from $1.8$\% to $2.4$\% and the ROE NEES from $2.1$ to $6.6$, showing that the inflation mechanism is needed for filter consistency.

It is clear from \autoref{tab:mc_breakdown_analysis} that each portion of the pipeline improves the accuracy, robustness, or consistency of the estimate, and none are redundant.

\subsection{Range observability with known maneuvers} \label{sec:range_results}

\hr{The same pipeline is run with the full six-element ROE on the maneuver schedule of \autoref{sec:scale_observability} for Herschel, over $20$ trials of three orbits. 
All 20 of them meet the specification. 
The ROE error falls to $2.2\%$ of the range, and $a \delta \lambda$ to $0.6$ m, once the first insertion burn is flown, against an unaligned error of $62\%$ without maneuvers. 
The range is therefore observable with this pipeline once suitable maneuvers are flown.}

\hr{\autoref{fig:mission_navigation} shows a representative trial, with the median attitude and ROE errors of the 20, where the ROE converges once the first insertion burn is flown. 
Its attitude error leaves the $3\sigma$ bound at times, reflecting the attitude over-confidence in \autoref{tab:mc_per_satellite}, while the angular-velocity bound is conservative.}

\subsection{Inference analysis}
The inference cost of each step of the MSCKF is given in \autoref{tab:inference_timing}, with the total for one filter update step in the bottom row.
Here only the time column is additive, with the per-stage memory numbers accounting only for the additional host and GPU memory each stage requires while it runs.
The bottom row on the other hand is the total the processor must provide, namely to fit the three networks and the filter state with the peak memory allocation for a single image pair. 
On the Orin, the CPU and GPU share a single memory pool so the RAM and VRAM columns show where each allocation is requested and the total is reported for the shared system. 
Some stages run more than once for every MSCKF filter step, in particular loop closure. 
From the Monte Carlo experiments, loop closure runs a median $2.03$ times per MSCKF filter step, and so this figure is adopted in \autoref{tab:inference_timing}. 
Note that since loop closure involves matching features, LightGlue matching is similarly increased from $1$ to $3.03$ calls per step. 

The largest single contributor to the pipeline cost is feature matching, accounting for half of the wall clock time for each filter step.
SON's timing and VRAM requirements are a function of the number of feature matches available. 
When performing odometry, the images are close in time and so a lot of feature matches are available. 
When performing loop closure fewer feature matches are available, and so this reduces the cost of the stage. 
Regardless, the total median cost of the MSCKF filter step is $1.05$ s, which is below the 5 s cadence used in this work.

\begin{table}[htbp]
    \centering
    \caption{Inference cost measured as compute time, host memory (RAM), and GPU memory (VRAM) for the MSCKF filter. Cost is measured per call.}
    \footnotesize
    \setlength\tabcolsep{2.5pt}%
\renewcommand{\arraystretch}{1.3}
\setlength{\heavyrulewidth}{0.4pt}
\setlength{\lightrulewidth}{0.4pt}
\setlength{\cmidrulewidth}{0.4pt}
\begin{tabular}{l r r r r}
\toprule
Stage & Calls/step & Cost $[ms]$ & RAM $[MB]$ & VRAM $[MB]$ \\
\midrule
SuperPoint & 1.00 & 113.3 & 308.9 & 304.8 \\
LightGlue & 3.03 & 172.2 & 272.6 & 303.3 \\
SON, odometry & 1.00 & 81.5 & 4.2 & 35.0 \\
SON, loop closure & 2.03 & 51.2 & 16.8 & 18.7 \\
MSCKF update & 1.00 & 229.0 & 287.7 & -- \\
\midrule
Filter step & & 1049 & 2518 & -- \\
\bottomrule
\end{tabular}
    \label{tab:inference_timing}
\end{table}

The network sizes vary based on the hyperparameters chosen in \autoref{tab:training_parameters}, but for the base case SuperPoint has $1.30$M parameters, LightGlue $11.85$M, and SON $14.54$M. 
This sums to a total $27.69$M parameters, which has a memory footprint of $110.8$ MB at a 32-bit float precision.

%%%%%%%%%%%%%%%%%%%%%%%%%%%%%%%%%%%%%%
\section{Conclusion}
%%%%%%%%%%%%%%%%%%%%%%%%%%%%%%%%%%%%%%
This work addresses a key challenge in monocular spacecraft navigation, that of unknown target pose estimation. 
First, a new dataset is developed, SPE3R+, which expands upon the previous work and adds more information, in the form of more images at higher resolution and metric depth, and more targets. 
This dataset is then used to fine-tune a state-of-the-art feature matcher, LightGlue, which has in the past only been used for terrestrial targets, and greatly enhances its performance. 
Then, spacecraft odometry is enabled by the development of the novel Spacecraft Odometry Network (SON), which greatly outperforms the previous baseline on SPE3R+ even when it uses the fine-tuned LightGlue. 
This is coupled with a Multi-State Constraint Kalman Filter (MSCKF), which tracks the direction of the Relative Orbit Elements (ROE), relative attitude, and relative angular velocity in a nominal state. 
This MSCKF uses the SON output as a measurement, as well as feature tracks that provide a local geometric constraint, and loop closure that uses SON again to estimate the odometry between the current image and a selected keyframe in the past to provide a global geometric constraint. 
These geometric constraints are required to avoid drift that is common with odometry measurements, since for attitude this is a form of dead-reckoning. 
\hr{The ROE are estimated up to scale, and a separate study shows that known orbit maneuvers make the range observable.} 

This pipeline is validated through 550 Monte Carlo experiments on 11 spacecraft which range from small satellites to large communications spacecraft and are never seen by the networks during training. 
This represents the true test scenario for this pipeline, as the targets are fully unknown. 
One target, SMART-1, is reported for transparency but excluded from the summary statistics, because its dark exterior leaves too little image texture.
\hr{After scale alignment, ROE errors remain below $4$\% of the range in $95\%$ of cases for every target.}
The median attitude error on the other hand depends only weakly on the distance to the target, and the attitude error is below \hr{$6.3^\circ$} in $95\%$ of cases.
Ablation studies validate both the SON architecture and the MSCKF design, showing the additional performance achieved through each of the design choices. 
The front-end is additionally validated on hardware-in-the-loop images from the SHIRT dataset, where SON maintains about $5$ times lower median translation unit direction error than the classical baseline, although its rotation error is slightly higher. 
The whole pipeline takes $1.05$ s per filter step on an NVIDIA AGX Orin, nearly $5$ times faster than the $5$ s image cadence.

In the future, the key assumptions outlined in \autoref{sec:training_and_implementation_details} will be removed or their effects quantified. 
In particular, the assumption of the camera always pointing at the target and that a reference-truth box is provided can be removed through the use of a small object detection network and estimator-based pointing control. 
The sim-to-real gap\hr{, which affects both rotation and translation direction,} must also be addressed, and it is likely that changes to the training of SON and the fine-tuning of LightGlue will be required to improve performance on real images. 
The compute of this overall pipeline would ideally also be reduced, and the overall design of LightGlue and SON must be scrutinized to reduce the network size while maintaining acceptable performance. 
\hr{Additional analysis is also needed to determine the effects of imperfect servicer knowledge and the accuracy needed for the initialization.}
Finally, inertia estimation will be pursued in future work to enable dynamical propagation without the need for so frequent measurements. 

%%%%%%%%%%%%%%%%%%%%%%%%%%%%%%%%%%%%%%
\acknowledgements
%%%%%%%%%%%%%%%%%%%%%%%%%%%%%%%%%%%%%%
This work was supported by Redwire Space Corporation through Stanford’s Center for AEroSpace Autonomy Research (CAESAR). 
The authors would like to thank Toby Bell and Emily Bates at Stanford University for their assistance in the development of the new SPE3R+ dataset, and general helpful discussions.
The authors would also like to thank Dr. Tae Ha Park at Nara Space for helpful discussions. 

The authors disclose the use of LLMs for general editing, paper formatting, and grammar corrections. 
The authors also disclose the use of LLMs for code assistance. 

%%%%%%%%%%%%%%%%%%%%%%%%%%%%%%%%%%%%%%
\bibliographystyle{IEEEtran}
\bibliography{references}

%%%%%%%%%%%%%%%%%%%%%%%%%%%%%%%%%%%%%%
\thebiography
%% This biostyle allows you to insert your photo size 1in X 1.25in
\begin{biographywithpic}
{Pol Francesch Huc}{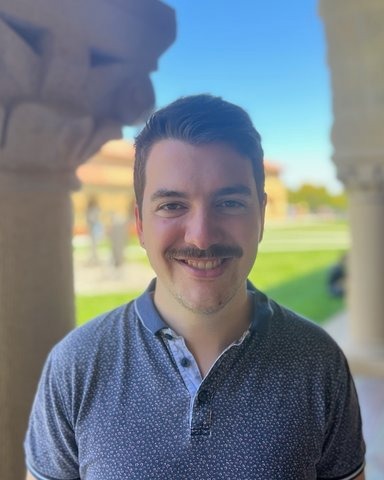}
is a Ph.D. student in the Space Rendezvous Lab. He received a Bachelor of Science in Aeronautical and Astronautical Engineering from Purdue University (2021), and a Master of Science in Aeronautics and Astronautics from Stanford University (2024). His research at Stanford focuses on vision-based navigation and characterization of resident space objects. Potential applications include space debris removal and refueling of defunct geostationary satellites with unprecedented autonomy and safety measures.
\end{biographywithpic} 

\begin{biographywithpic}
{Simone D'Amico}{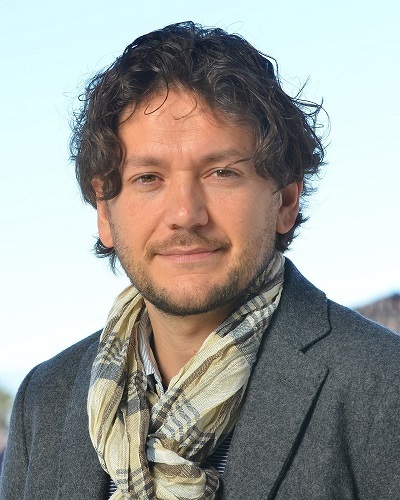}
is Associate Professor of Aeronautics and Astronautics at Stanford University, Science Fellow at the Hoover Institution, Co-Founder and Chief Science Officer at EraDrive Inc. He is the Director of the Space Rendezvous Laboratory, and Co-Director of the Center for AEroSpace Autonomy Research (CAESAR). He has 23+ years of experience in research and development of autonomous spacecraft and distributed space systems. He developed and deployed the distributed Guidance, Navigation, and Control (GNC) system of several formation-flying, swarming, rendezvous and proximity operations missions. Currently, he is the institutional PI of four autonomous satellite swarms funded by NASA and NSF. Dr. D’Amico is Fellow of AIAA and AAS. He is on the Advisory Board of four space start-ups focusing on future applications of distributed space systems. He was the recipient of several awards, most recently the 2024 NASA Ames Honor Award and the 2025 SpaceNews Icon Award for the Starling mission, Best Paper Awards at AMOS (2025), IAF (2022), IEEE (2021), AIAA (2021), AAS (2019) conferences, and the M. Barry Carlton Award by IEEE (2020). He received the B.S. and M.S. degrees from Politecnico di Milano (2003) and the Ph.D. degree from Delft University of Technology (2010).
\end{biographywithpic}

\end{document}